\documentclass[10pt,journal,compsoc]{IEEEtran}

\ifCLASSOPTIONcompsoc
  \usepackage[nocompress]{cite}
\else
  \usepackage{cite}
\fi

\usepackage[colorlinks,linkcolor=red]{hyperref}
\usepackage{url}            %
\usepackage{booktabs}       %
\usepackage{amsfonts}       %
\usepackage{nicefrac}       %
\usepackage{microtype}      %
\usepackage{xcolor}         %
\usepackage{graphicx}
\usepackage{wrapfig}
\usepackage{multirow}
\usepackage{graphicx}
\usepackage{float}
\usepackage{amsmath}
\usepackage{bbding}
\usepackage{color}

\newcommand{\md}[1]{\textcolor[rgb]{0.0,0.0,0.0}{#1}}

\ifCLASSINFOpdf
\else
\fi

\begin{document}

\title{CAP-UDF: Learning Unsigned Distance Functions Progressively from Raw Point Clouds with Consistency-Aware Field Optimization}

\author{Junsheng Zhou\IEEEauthorrefmark{1},
        Baorui Ma\IEEEauthorrefmark{1}, 
        Shujuan Li,
        Yu-Shen Liu,~\IEEEmembership{Member,~IEEE,} Yi Fang, Zhizhong Han
\IEEEcompsocitemizethanks{
\IEEEcompsocthanksitem Junsheng Zhou, Baorui Ma and Shujuan Li are with the School of Software, Tsinghua University, Beijing, China. E-mail: zhoujs21, mbr18, lisj22@mails.tsinghua.edu.cn
\IEEEcompsocthanksitem Yu-Shen Liu is with the School of Software, Tsinghua University, Beijing, China. E-mail: liuyushen@tsinghua.edu.cn
\IEEEcompsocthanksitem Yi Fang is with Center for Artificial Intelligence and Robotics, New York University Abu Dhabi, Abu Dhabi, UAE. E-mail: yfang@nyu.edu
\IEEEcompsocthanksitem Zhizhong Han is with the Department of Computer Science, Wayne State University, USA E-mail: h312h@wayne.edu
}
\thanks{Junsheng Zhou and Baorui Ma contribute equally to this work. The corresponding author is Yu-Shen Liu. This work was supported by National Key R\&D Program of China (2022YFC3800600), and the National Natural Science Foundation of China (62272263, 62072268). Project page: \url{https://junshengzhou.github.io/CAP-UDF}.}
}


\IEEEtitleabstractindextext{%
\begin{abstract}

Surface reconstruction for point clouds is an important task in 3D computer vision. Most of the latest methods resolve this problem by learning signed distance functions from point clouds, which are limited to reconstructing closed surfaces. Some other methods tried to represent open surfaces using unsigned distance functions (UDF) which are learned from ground truth distances. However, the learned UDF is hard to provide smooth distance fields due to the discontinuous character of point clouds. In this paper, we propose CAP-UDF, a novel method to learn consistency-aware UDF from raw point clouds. We achieve this by learning to move queries onto the surface with a field consistency constraint, where we also enable to progressively estimate a more accurate surface. Specifically, we train a neural network to gradually infer the relationship between queries and the approximated surface by searching for the moving target of queries in a dynamic way. Meanwhile, we introduce a polygonization algorithm to extract surfaces using the gradients of the learned UDF. We conduct comprehensive experiments in surface reconstruction for point clouds, real scans or depth maps, and further explore our performance in unsupervised point normal estimation, which demonstrate non-trivial improvements of CAP-UDF over the state-of-the-art methods.

\end{abstract}

}

\maketitle

\IEEEdisplaynontitleabstractindextext

\IEEEpeerreviewmaketitle

\section{Introduction}
Reconstructing surfaces from 3D point clouds is vital in 3D vision, robotics and graphics. It bridges the gap between raw point clouds that can be captured by 3D sensors and the editable surfaces for various downstream applications. Recently, Neural Implicit Functions (NIFs) have achieved promising results by training deep networks to learn Signed Distance Functions (SDFs) \cite{park2019deepsdf, jiang2020local, michalkiewicz2019deep, duan2020curriculum} or occupancies \cite{mescheder2019occupancy, peng2020convolutional, mi2020ssrnet, chen2019learning}. With the learned NIFs, we can extract a polygon mesh as a continuous iso-surface of a discrete scalar field using the marching cubes algorithm \cite{lorensen1987marching}. However, the NIFs approaches based on learning internal and external relations can only reconstruct closed surfaces. The limitation prevents NIFs from representing most real-world objects such as cars with inner structures, clothes with unsealed ends or 3D scenes with open walls and holes.

As a remedy, state-of-the-art methods \cite{chibane2020neural,zhao2021learning,venkatesh2021deep} learn Unsigned Distance Functions (UDFs) as a more general representation to reconstruct surfaces from point clouds. However, these methods can not learn UDFs with smooth distance fields near surfaces, due to the discontinuous character of point clouds, even using ground truth distance values or large scale meshes during training. Moreover, most UDF approaches failed to extract surfaces directly from unsigned distance fields. Particularly, they rely on post-processing such as generating dense point clouds from the learned UDFs for Ball-Pivoting-Algorithm (BPA) \cite{bernardini1999ball} to extract surfaces, which is very time-consuming and also leads to surfaces with discontinuity and low quality.

To solve these issues, we propose a novel method named CAP-UDF to learn \textbf{C}onsistency-\textbf{A}ware UDFs \textbf{P}rogressively from raw point clouds. We learn to move 3D queries to reach the approximated surface progressively with a field consistency constraint, and introduce a polygonization algorithm to extract surfaces from the learned UDFs from a new perspective. Our method can learn UDFs from a single point cloud without requiring ground truth unsigned distances, point normals, or a large scale training set. Specifically, given queries sampled in 3D space as input, we learn to move them to the approximated surface according to the predicted unsigned distances and the gradient at the query locations. More appealing solutions \cite{atzmon2020sal,gropp2020implicit,atzmon2020sald,ma2021neural} have been proposed to learn SDFs from raw point clouds by inferring the relative locations of a query and its closest point in the point cloud. However, since the raw point cloud is a highly discrete approximation of the surface, the closest point of the query is always inaccurate and ambiguous, which makes the network difficult to converge to an accurate UDF due to the inconsistent or even conflicting optimization directions in the distance field.

Therefore, in order to encourage the network to learn a consistency-aware and accurate unsigned distance field, we propose to dynamically search the optimization target with a specially designed loss function constraining the consistency in the field.
We also progressively infer the mapping between queries and the approximated zero iso-surface by using well-moved queries as additional priors for promoting further convergence.
To extract a surface in a direct way, we propose to use the gradient of the learned UDFs to determine whether two queries are on the same side of the approximated surface or not.
In contrast to NDF \cite{chibane2020neural} which also learns UDFs but outputs dense point clouds for BPA \cite{bernardini1999ball} to generate meshes, our method shows great advantages in efficiency and accuracy due to the straightforward surface extraction.

With the ability of learning a consistency-aware and accurate unsigned distance field, we make a step forward and extend our method \cite{Zhou2022CAP-UDF} for unsupervised point normal estimation, where our superior performance is demonstrated by both quantitative and qualitative results with the state-of-the-art unsupervised and supervised normal estimation methods. Furthermore, we evaluate the performance of our method using point clouds from depth sensors and show great advantages over the state-of-the-art NeRF-based \cite{mildenhall2020nerf} or TSDF-based \cite{azinovic2022neural, wang2022go} methods using RGB-D images, where we only take depth maps as input without requiring colored images.

Our main contributions can be summarized as follows.
\begin{itemize}
    \item We propose a novel neural network named CAP-UDF that learns consistent-aware UDFs from raw point clouds without requiring ground truth distance values or point normals. Our method gradually infers the relationship between 3D query locations and the approximated surface with a field consistent loss. 
    \item We introduce an algorithm for directly extracting high-fidelity iso-surfaces with arbitrary topology using the gradients of the learned UDFs.
    \item We conduct comprehensive experiments in surface reconstruction for synthetic point clouds, real scans or depth maps, and further explore our performance in the unsupervised point normal estimation task. The experimental results demonstrate our significant improvements over the state-of-the-art methods under the widely used benchmarks. 

\end{itemize}

\section{Related Works}
Surface reconstruction from 3D point clouds has been studied for decades. Classic optimization-based methods \cite{edelsbrunner1994three, bernardini1999ball, kazhdan2006poisson, kazhdan2013screened} tried to resolve this problem by inferring continuous surfaces from the geometry of point clouds. With the rapid development of deep learning \cite{zhou2023uni3d, liu2022spu, wen20223d, Jiang2019SDFDiffDRcvpr, hutaoaaai2020, zhou2022-3DOAE, wen2022pmp, xiang2021snowflake, Zhou2023VP2P}, the neural networks have shown great potential in reconstructing 3D surfaces \cite{li2022learning, LPI, chen2021unsupervised, han2020reconstructing, wang2021neus, oechsle2021unisurf, lindell2022bacon, wang2022improved, Zhou2022CAP-UDF, huang2023neusurf, zhou2023levelset, jin2023multi, ma2023geodream,ma2023towards}. In the following, we will briefly review the studies of deep learning based methods.

\subsection{Neural Implicit Surface Reconstruction}
In the past few years, a lot of advances have been made in 3D surface reconstruction with Neural Implicit Functions (NIFs). The NIFs approaches \cite{chabra2020deep, peng2020convolutional, mescheder2019occupancy, jiang2020local, liu2021deep, mi2020ssrnet, lombardi2020scalable, park2019deepsdf, martel2021acorn} use either binary occupancies \cite{mescheder2019occupancy, peng2020convolutional, mi2020ssrnet, chen2019learning} or signed distance functions (SDFs) \cite{park2019deepsdf, jiang2020local, michalkiewicz2019deep, duan2020curriculum} to represent 3D shapes or scenes, and then use the marching cubes \cite{lorensen1987marching} algorithm to reconstruct surfaces from the learned implicit functions. 
Earlier studies \cite{mescheder2019occupancy, park2019deepsdf, chen2019learning} use an encoder \cite{mescheder2019occupancy, chen2019learning} or an optimization based method \cite{park2019deepsdf} to embed the shape into a global latent code, and then use a decoder to reconstruct the shape. To obtain more detailed geometry, some methods \cite{genova2020local, genova2019learning, tretschk2020patchnets, chabra2020deep, chibane2020implicit, jiang2020local, mi2020ssrnet, PredictiveContextPriors, On-SurfacePriors} proposed to leverage more latent codes to capture local shape priors. To achieve this, the point cloud is first split into different uniform grids \cite{chabra2020deep, chibane2020implicit, jiang2020local} or local patches \cite{genova2020local, genova2019learning, tretschk2020patchnets}, and a neural network is then used to extract a latent code for each grid/patch. Some recent methods propose to learn NIFs from a new perspective, such as implicit moving least-squares surfaces \cite{liu2021deep}, differentiable poisson solver \cite{peng2021shape}, iso-points \cite{yifan2021iso} , point convolution \cite{boulch2022poco} or predictive context learning \cite{PredictiveContextPriors}. However, the NIFs approaches can only represent closed shapes due to the characters of occupancies and SDFs.

\subsection{Learning Unsigned Distance Functions}
To model general shapes with open and multi-layer surfaces, NDF \cite{chibane2020neural} learns unsigned distance functions to represent shapes by predicting an unsigned distance from a query location to the continuous surface. However, NDF merely predicts dense point clouds as the output, which requires time-consuming post-processing for mesh generation and also struggles to retain high-quality details of shapes. In contrast, our method is able to extract surfaces directly from the gradients of the learned UDFs. Following studies use image features \cite{zhao2021learning} or query side relations \cite{ye2022gifs} as additional constraints to improve reconstruction accuracy, some other works advance UDFs for normal estimation \cite{venkatesh2021deep} or semantic segmentation \cite{wang2022rangeudf}. However, these methods require ground truth unsigned distances or even a large scale meshes during training, which makes it hard to provide smooth distance fields near the surface due to the discontinuous character of point clouds. While our method does not require any additional supervision but raw point clouds during training, which allows us to reconstruct surfaces for real point cloud scans. In a differential manner, a concurrent work named MeshUDF \cite{guillard2021meshudf} meshes UDFs from the dynamic gradients during training using a voting schema. On the contrary, we learn a consistency-aware UDF first and extract the surface from stable gradients during testing. Moreover, our surface extraction algorithm is simpler to use, which is implemented based on the marching cube algorithm.

\begin{figure*}[!t]
  \centering
  \includegraphics[width=2\columnwidth]{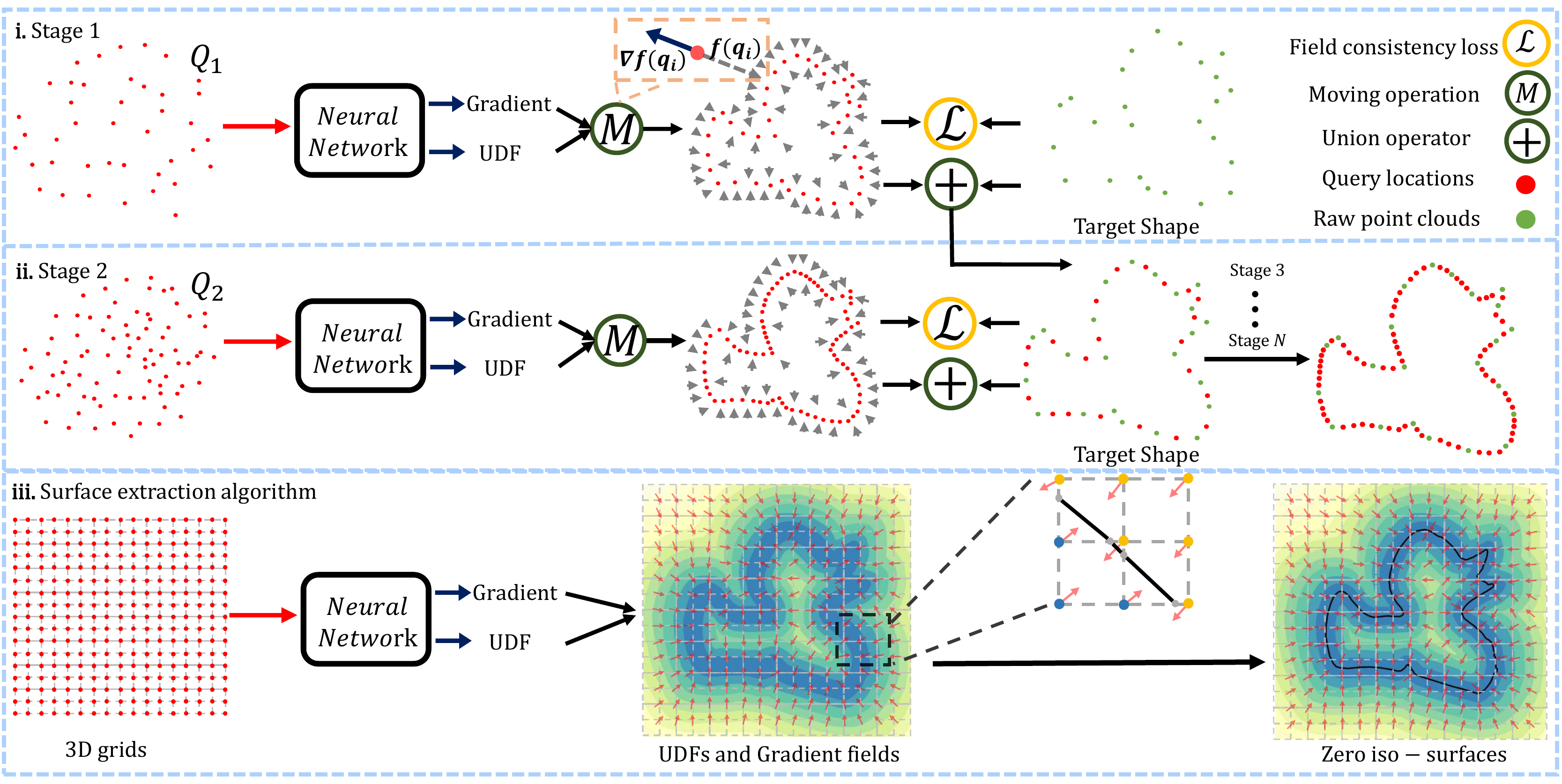}
  \caption{Overview of CAP-UDF. Given a 3D query $q_i \in {Q_1}$ as input, the neural network predicts the unsigned distance $f(q_i)$ of $q_i$ and moves $q_i$ against the direction of gradient at $q_i$ with a stride of $f(q_i)$. The field consistency loss is then computed between the moved queries $q'_i$ and the target point cloud $P$ as the optimization target. After the network converges in the current stage, we update $P$ with a subset of $q'_i$ as additional priors to learn more local details in the next stage. Finally, we use the gradients of the learned UDFs in the field to model the relationship between different 3D grids and extract iso-surfaces. }
  \label{fig:overview}
  \vspace{-0.5cm}
\end{figure*}

\subsection{Surface Reconstruction from Raw Point Clouds}
Learning implicit functions directly from raw point clouds without ground truth signed/unsigned distances or occupancy labels is more challenge.  Current studies introduce sign agnostic learning with a specially designed network initialization \cite{atzmon2020sal}, constraints on gradients \cite{atzmon2020sald} or geometric regularization \cite{gropp2020implicit} for learning SDFs from raw data. Neural-Pull \cite{ma2021neural} uses a new way of learning SDFs by pulling nearby space onto the surface. However, they aim to learn signed distances and hence can not reconstruct complex shapes with open or multi-layer surfaces. In contrast, our method is able to learn a continuous unsigned distance function from point clouds, which allows us to reconstruct surfaces for shapes and scenes with arbitrary typology.

\section{Method}

In this section, we present CAP-UDF, a novel framework to learn consistency-aware UDFs progressively from raw point clouds. We introduce the schema of learning UDFs from raw point clouds in Sec. \ref{sec:3.1} with the consistency-aware field optimization shown in Sec. \ref{sec:3.2}. We propose the progressive surface approximation strategy in Sec. \ref{sec.progress}. The gradient-based surface extraction algorithm for UDFs is described in Sec. \ref{sec:3.4}. We further extend CAP-UDF to unsupervised point normal estimation in Sec. \ref{sec:3.5}. The overview of CAP-UDF is shown in Fig. \ref{fig:overview}.

\textbf{Method overview.} We design a neural network to learn UDFs that represent 3D shapes and scenes. Given a 3D query location $q = [x,y,z]$, a learned UDF $f$ predicts the unsigned distance value $s = f(q) \in \mathbb{R}$. Current methods rely on ground truth distance values generated from continuous surfaces and employ a neural network to learn $f$ as a regression problem. Different from these methods, we aim to learn $f$ from a raw point cloud $P = \{p_i, i \in [1,N]\}$ without using ground truth unsigned distances. Furthermore, these methods require post-processing \cite{chibane2020neural} or additional supervision \cite{ye2022gifs} to generate meshes. On the contrary, we introduce an algorithm to extract surfaces directly from $f$ using the gradient field $\nabla f$. 

\begin{figure}[!t]
    \centering
    \includegraphics[width=0.41\textwidth]{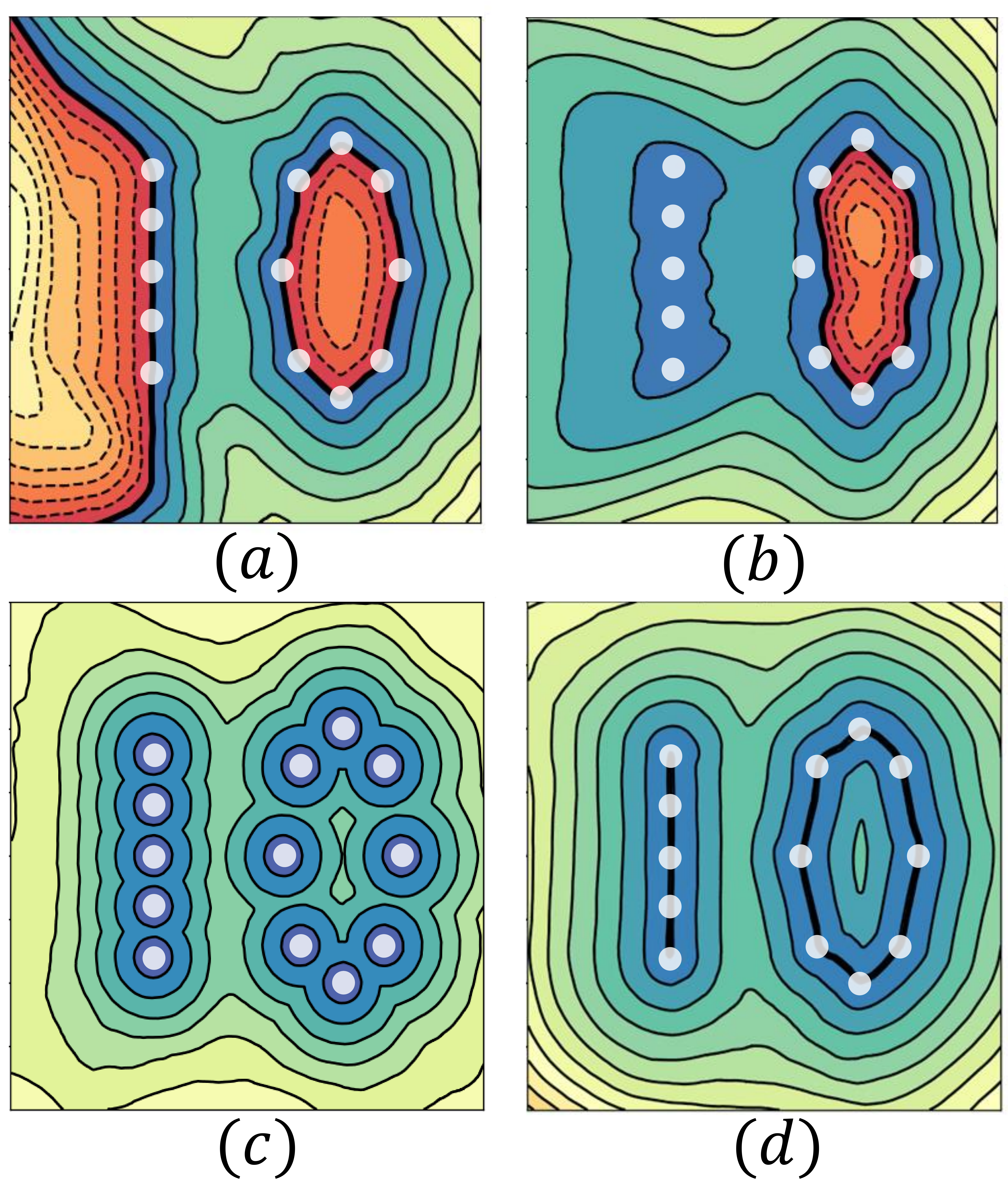}
    \caption{The level-sets show the distance fields learned by (a) Neural-Pull, (b) SAL, (c) NDF, and (d) Ours. The color of blue and red represent positive and negative distance, respectively. The darker the color, the closer it is to the approximated surface.}
    \label{fig:levelset}

\end{figure}

\subsection{Learn UDFs from Raw Point Clouds}
\label{sec:3.1}
We introduce a novel neural network to learn a continuous UDF $f$ from a raw point cloud. We demonstrate our idea using a 2D point cloud $S$ in Fig. \ref{fig:overview}(a), where $S$ indicates some discrete points of a continuous surface. Specifically, given a set of query locations $Q=\{q_i, i \in [1, M]\}$ which is randomly sampled around $S$, the network moves $q_i$ against the direction of the gradient $g_i$ at $q_i$ with a stride of predicted unsigned distance value $f(q_i)$. The gradient $g_i$ is a vector that presents the partial derivative of $f$ at $q_i=[x_i, y_i, z_i]$, which can be formulated as $g_i = \nabla f(q_i) = [{\partial f} / {\partial x}, {\partial f} / {\partial y}, {\partial f} / {\partial z}] $. $g_i$ indicates the direction of the greatest unsigned distance change in 3D space, which points the direction away from the surface, therefore moving $q_i$ against the direction of $g_i$ will find a path to the surface of $S$. The moving operation can be formulated as:
\begin{equation}
    \label{eq:move}
    z_i = q_i - f(q_i) \times \nabla f(q_i)/||\nabla f(q_i)||_2,
\end{equation}

\noindent where $z_i$ is the location of the moved query $q_i$, and $\nabla f(q_i)/||\nabla f(q_i)||_2$ is the normalized gradient $g_i$, which indicates the direction of $g_i$. The moving operation is differentiable in both the unsigned distance value and the gradient, which allows us to optimize them simultaneously during training.

The four examples in Fig. \ref{fig:levelset} show the distance fields learned by Neural-Pull \cite{ma2021neural}, SAL \cite{atzmon2020sal}, NDF \cite{chibane2020neural} and our method for a sparse 2D point cloud $P$ which only contains 13 points.
One main branch to learn signed or unsigned distance functions for point clouds is to directly minimize the mean squared error between the predicted distance value $f(q_i)$ and the \md{Euclidean distance} between $q_i$ and its nearest neighbour in $P$, as proposed in NDF and SAL. However, as shown in Fig. \ref{fig:levelset}(c), NDF leads to an extremely discrete distance field. To learn a continuous distance field, NDF introduces ground truth distance values extracted from the continuous surface as extra supervision, which prevents it from learning from raw point clouds. SAL shows a great capacity in learning SDFs for watertight shapes using a carefully designed initialization. However, as shown in Fig. \ref{fig:levelset}(b), SAL fails to converge to a multi-part structure since the network is initialized as a single layer shape prior. Neural-Pull uses a similar way as ours to pull queries onto the surface, thus also learns a continuous signed distance field as shown in Fig. \ref{fig:levelset}(a). However, the nature of SDF prevents Neural-Pull from reconstructing open surfaces like the $``1"$ on the left of Fig. \ref{fig:levelset}(a). As shown in Fig. \ref{fig:levelset}(d), our method can learn a continuous level set of distance field and can also represent open surfaces.

One way to extend Neural-Pull directly to learn UDFs is to predict a positive distance value for each query and pull it to the nearest neighbour in $P$.
However, for shapes with complex topology, this optimization is often ambiguous due to the discontinuous character of raw point clouds. Hence, we resolve this problem by introducing consistency-aware field learning.

\begin{figure}[tb]
    \centering
    \includegraphics[width=0.48\textwidth]{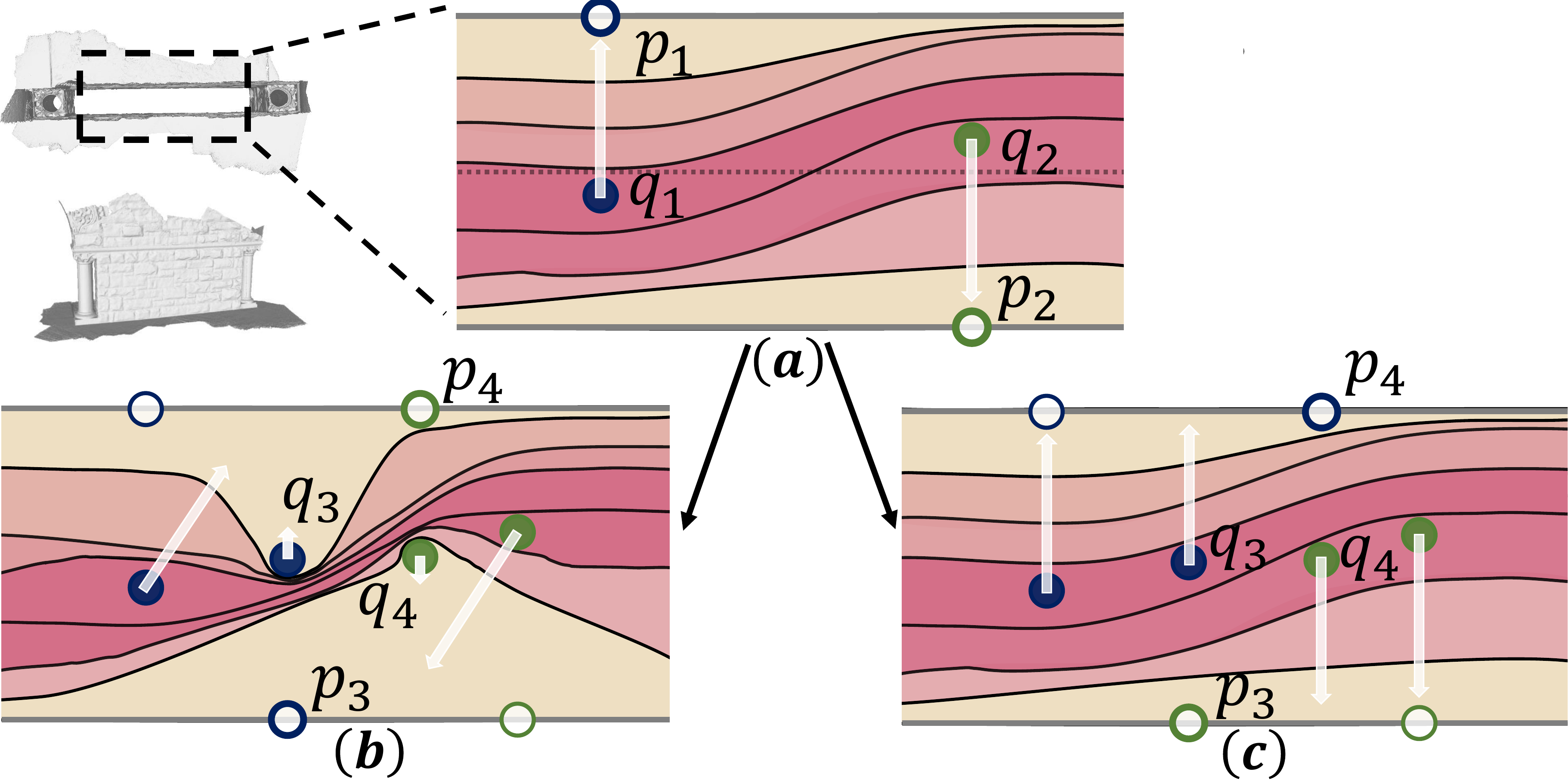}
    \caption{\md{Illustration of optimizing with different losses. (a) The initial distance field. (b) The distorted field with local minimum caused by inconsistent optimization with naive loss in Eq. (\ref{eq:l2loss}). (c) Optimizing with our consistency-aware loss in Eq. (\ref{eq:gcloss}) leads to correct and continuous distance field. }  }
    \label{fig:loss}
\end{figure}

\subsection{Consistency-Aware Field Learning}
\label{sec:3.2}
Neural-Pull leverages a mean squared error to minimize the distance between the moved query $z_i$ and the nearest neighbour $n_i$ of $q_i$ in $P$:
\begin{equation}
    \label{eq:l2loss}
    \mathcal{L} = \frac{1}{M} \sum_{i \in [1,M]} {||z_i-n_i||_2^2}.
\end{equation}

However, the direct optimization of the loss in Eq. (\ref{eq:l2loss}) will form a distorted field and lead some queries to get stuck in the local areas due to the conflict optimization which makes the network difficult to converge. \md{We show a 2D demonstration of learning UDFs for a double-deck wall using the loss Eq. (\ref{eq:l2loss}) in Fig. \ref{fig:loss}(b).}
Given $p_1$ and $p_2$ as two discrete points in two different decks of the wall, $q_1$ and $q_2$ are two queries whose closest neighbours are $p_1$ and $p_2$, respectively. Optimizing the network using $q_1$ and $q_2$ by minimizing Eq. (\ref{eq:l2loss}) or our proposed loss in Eq. (\ref{eq:gcloss}) will lead to an unsigned distance field as in Fig. \ref{fig:loss}(a). 
Assuming in the next training batch, $q_3$ and $q_4$ are two queries whose closest neighbours are $p_3$ and $p_4$. If we use the loss in Eq. (\ref{eq:l2loss}), the optimization target of $q_3$ is to minimize $ \mathcal{L} = {||z_3-p_3||_2^2}$. Notice that the target point $p_3$ is located on the lower surface, however the opposite direction of gradient around $q_3$ is upward at this moment. Therefore, the partial derivative ${\partial \mathcal{L}} / {\partial z_3}$ leads to a decrease in the unsigned distance value $f(q_3)$ predicted by the network. The case of $q_4$ is optimized similarly. An immediate consequence is that the inconsistent optimization directions will form a distorted fields that has local minima of unsigned distance values at $q_3$ and $q_4$ as in Fig. \ref{fig:loss}(b). However, this situation causes other query points around point $q_3$ or $q_4$ to get stuck in the distorted fields and unable to move to the correct location, thus making the network hard to converge.

\begin{figure}[tb]
    \centering
    \includegraphics[width=0.45\textwidth]{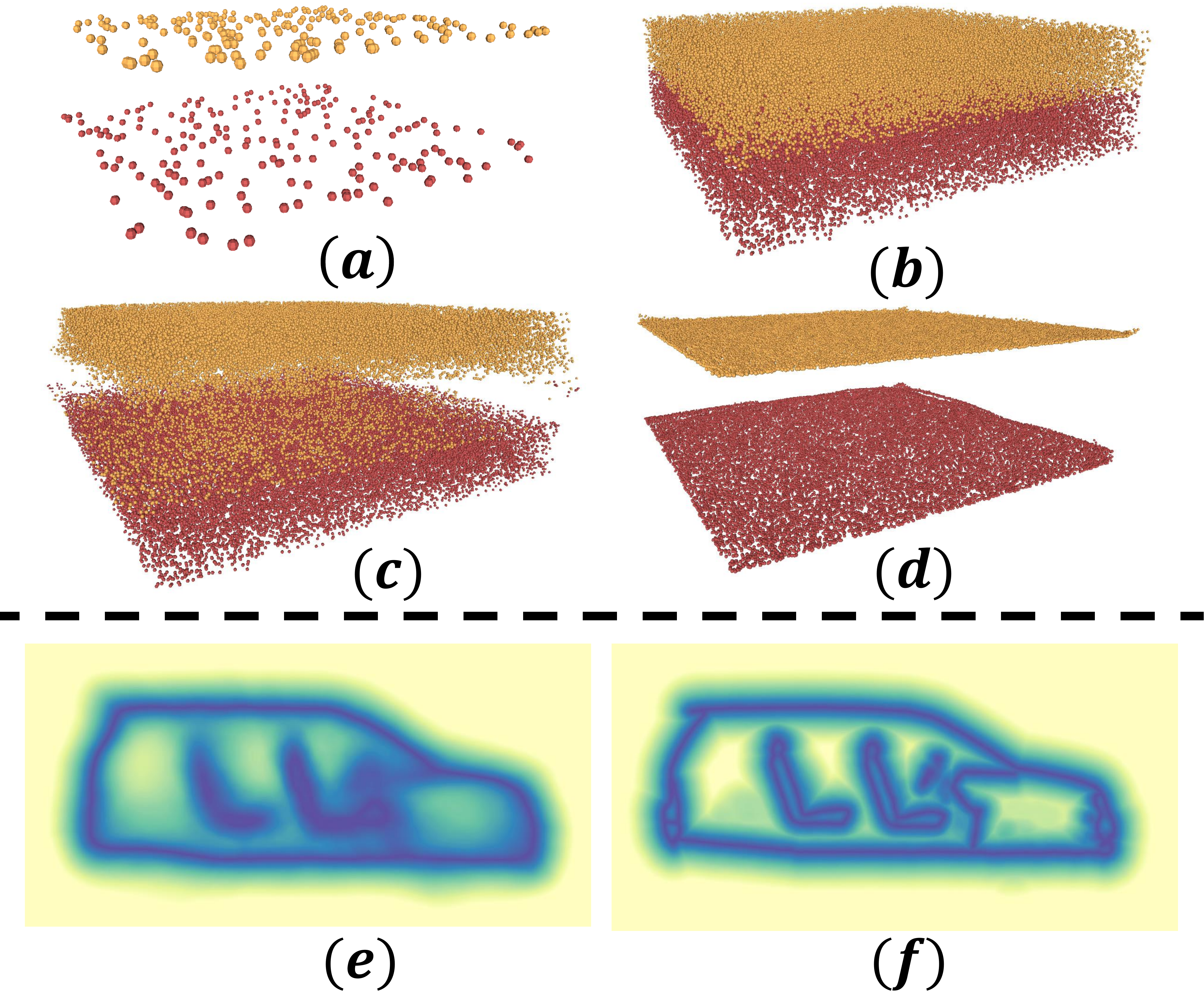}
    \vspace{-0.2cm}
    \caption{\md{Demonstration experiments on the effectiveness of consistency-aware loss. (a) The input point cloud of a double-deck wall. (b) The randomly sampled query points between two decks. (c, d) The moved queries optimized by naive loss in Eq. (\ref{eq:l2loss}) and our loss in Eq. (\ref{eq:gcloss}). (e,f) The learned distance field of a car with inner structure by the loss in Eq. (\ref{eq:l2loss}) and our loss in Eq. (\ref{eq:gcloss}).} }
    \label{fig:loss_exp}
\end{figure}

\begin{figure*}[h]
  \centering
  \includegraphics[width=2\columnwidth]{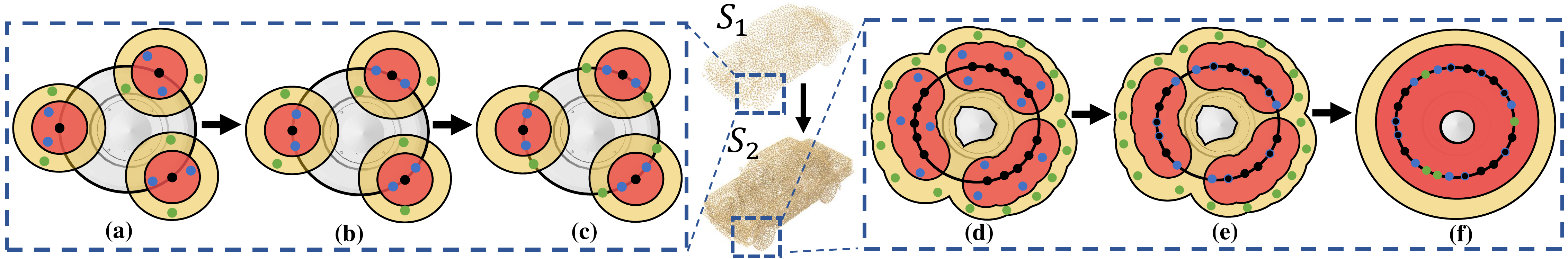}
  \vspace{-0.3cm}
  \caption{\md{Illustration of progressively approximating the surface. (a) We sample queries on the high confidence (red) and low confidence (yellow) regions. (b, c) The optimization process in the first stage for learning UDFs by moving queries as Eq. (\ref{eq:move}). (d) We update the raw points with the moved queries in both regions as additional priors for learning more local details in the next stage. (e,f) More continuous and accurate field is achieved with the progressive learning schema.}}
  \label{fig:multistep}
\end{figure*}

To address this issue, we propose a loss function which can keep the consistency of unsigned distance fields to avoid the conflicting optimization directions.
Specifically, instead of strictly constraining the convergence target before forward propagation as Eq. (\ref{eq:l2loss}), we first predict the moving path of a query location $q_i$ and move it using Eq. (\ref{eq:move}) to $z_i$, then look for the surface point $p_i$ in $P$ which is the closest to $z_i$ and minimize the distance between $z_i$ and $p_i$. As shown in Fig. \ref{eq:gcloss}(c), after moving $q_3$ against the gradient direction with a stride of $f(q_3)$ to $z_3$, the closest surface point of $z_3$ lies on the upper deck, so the distance fields remain continuous and are optimized correctly.
In practice, we can achieve this by using the \md{Chamfer distance} as a loss function, formulated as:
\begin{equation}\small
    \label{eq:gcloss}
    \begin{split}
         \mathcal{L}_{\rm CD} = \frac{1}{M} \sum_{i \in [1,M]}  &\mathop{min}\limits_{j \in [1, N]}{||z_i-p_j||_2} 
        \\&+ \frac{1}{N} \sum_{j \in [1,N]} \mathop{min}\limits_{i \in [1, M]}{||p_j-z_i||_2},
    \end{split}
\end{equation}
where $z_i, i\in [1, M]$ are the moved queries and $p_j, j\in [1,N]$ indicate the raw point cloud. 
We also use toy examples as shown in Fig. \ref{fig:loss_exp} to show the advantage of our proposed field consistency loss. We learn UDFs for a raw point cloud of a double-deck wall as shown in Fig. \ref{fig:loss_exp}(a).  Fig. \ref{fig:loss_exp}(b) shows the randomly sampled query locations between the two decks of the wall where the different colors distinguish the queries that are closer to the upper or lower deck of the wall. Fig. \ref{fig:loss_exp}(c) and Fig. \ref{fig:loss_exp}(d) indicate the moved queries by loss in Eq. (\ref{eq:l2loss}) and Eq. (\ref{eq:gcloss}). It can be seen that our proposed loss can move most of the queries to the correct surface position, while the Neural-Pull loss can not move queries in many places or moves queries to the wrong places due to the field inconsistency in optimization. Fig. \ref{fig:loss_exp}(e) and Fig. \ref{fig:loss_exp}(f) show the learned distance field of a car with inner structure by the loss in Eq. (\ref{eq:l2loss}) and our loss, respectively.

\subsection{Progressive Surface Approximation}
\label{sec.progress}
Moreover, in order to predict unsigned distance values more accurately and learn more local details, we propose a progressive learning strategy by taking the intermediate results of moved queries as additional priors. Given a raw point cloud which is a discrete representation of the surface, 
we have made a reasonable assumption: the closer the query location is to the given point cloud, the smaller the error of searching the target point on the given point cloud. \md{We provide the proof of this assumption in Sec. 1 of the supplementary}. Based on this assumption, we set up two regions: the high confidence region with small errors and the low confidence region with large errors. We sample query points in the high confidence region to help train the network and sample auxiliary points in the low confidence region to move onto the estimated surface by network gradient after network convergence at current stage, where the moved auxiliary points are regarded as the surface prior for the next stage. Notably, the auxiliary points do not participate in network training since these points with low confidence will lead to a large error and affect network training. Since the low confidence regions which are not optimized explicitly during training are distributed interspersed between the high confidence regions, according to the integral Monotone Convergence Theorem \cite{bibby1974axiomatisations}, the UDFs and gradients predicted by the low confidence region are a smooth expression of the trained high confidence region. We use the moved queries and auxiliary points to update $S$. According to the updated point cloud, we re-divide the regions with high confidence and low confidence and re-sample the query points and auxiliary points for the next stage.

We demonstrate our idea using a 2D case in Fig. \ref{fig:multistep}.(a) We divided the regions with high confidence (red region) and low confidence (yellow region) based on the given raw point cloud $S _{1}$ (black dots) and then sample query points $Q_{1}=\{q_i, i \in [1, M]\}$ (blue dots) and auxiliary points (green dots) $A_{1}=\{a_i, i \in [1, M]\}$. (b) We train the network to learn UDFs by moving the query locations $Q_{1}$ using Eq. (\ref{eq:move}), and optimize the network by minimizing Eq. (\ref{eq:gcloss}). (c) After the network convergence at current stage, we move query points $Q$ and auxiliary points $A$ to the estimated surface position by the gradient of network, $S_{1}^{'} = p - f(p) \times \nabla f(p)/||\nabla f(p)||_2, p \subset  Q _{1} \cup A_{1}$. (d) We use the moved points $S_{1}^{'}$ to update $S$, $S _{2}= S _{1} \cup S_{1}^{'}$. According to the updated point cloud $S _{2}$, we re-divided the regions with high confidence and low confidence and re-sampled the query points $Q_{2}$ and auxiliary points $A_{2}$. (e) We continue to train the network by moving query points $Q_{2}$ to the updated $S_{2}$, and then update $S$ by combining the moved $Q_2$ and $A_2$ with $S_2$. (f) Because of the more continuous surface, the network can leverage a prior information to learn more accurate UDFs with more local details of shapes.

\subsection{Surface Extraction Algorithm}
\label{sec:3.4}
\begin{figure}[h]
    \centering
    \includegraphics[width=0.45\textwidth]{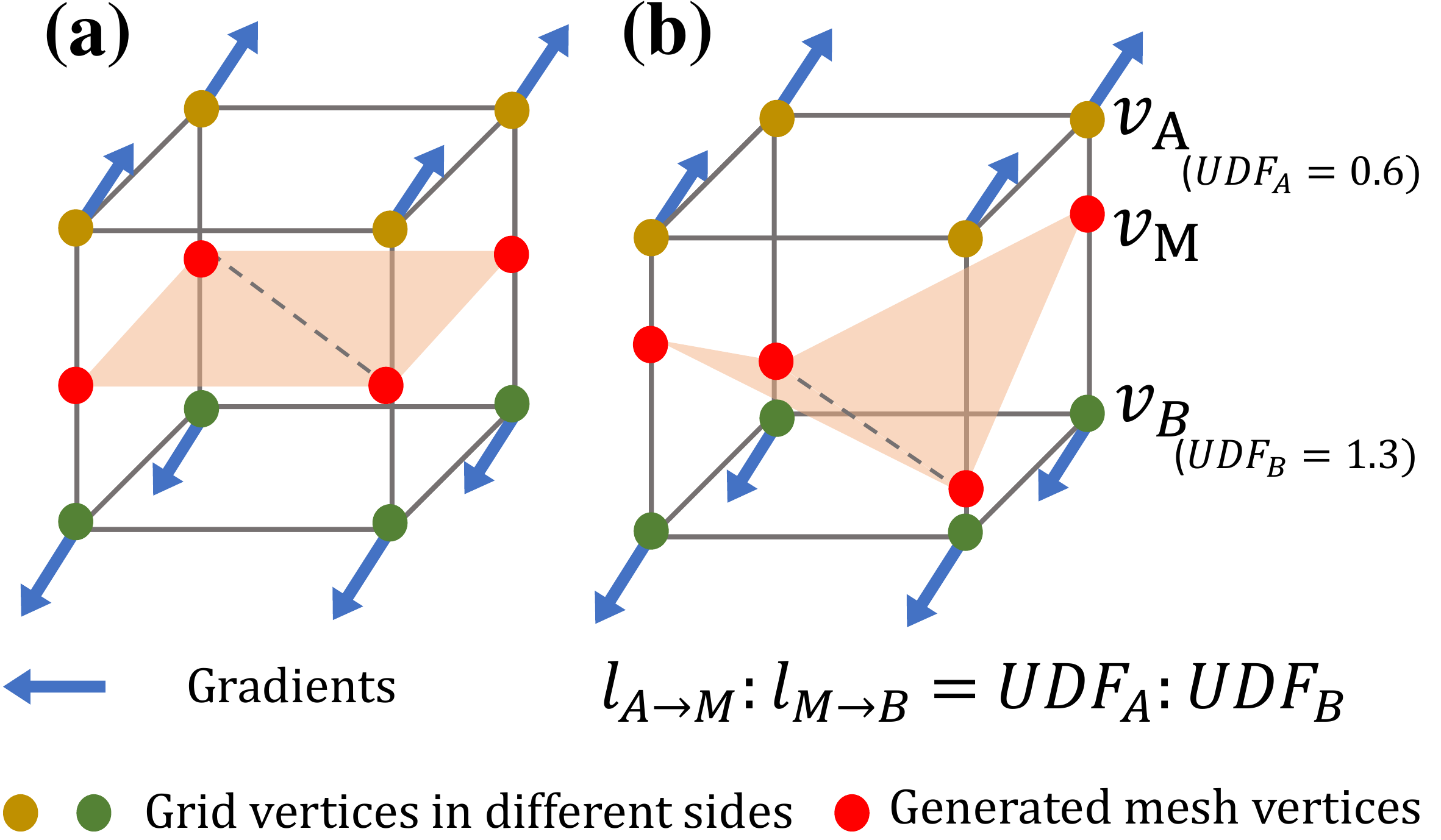}
    \caption{Surface extraction algorithm.}
    \label{fig:mcubes}
\end{figure}

\noindent Unlike SDFs, UDFs fail to  extract surfaces using the marching cubes since UDFs cannot perform inside/outside tests on 3D grids. To address this issue, we propose to use the gradient field $\nabla f$ to determine whether two 3D grid locations are on the same side or the opposite side of the surface approximated by the point clouds $P$.
We make an assumption that the space can be divided into two sides on a micro-scale of the surface, where the 3D query locations of different sides denoted as $Q_{in} = \{{q_{in}}^i, i \in [1,L]\}$ and $Q_{out} = \{{q_{out}}^i, i \in [1,I]\}$. For two queries ${q_{in}}^i$ and ${q_{out}}^j$ in different sides of the surface, the included angle between the directions of the gradients  $\nabla f({q_{in}}^i)$ and $\nabla f({q_{out}}^j)$ are more than 90 degrees, which can be formulated as $\nabla f({q_{in}}^i) \cdot \nabla f({q_{out}}^j) < 0$. On the contrary, for two queries ${q_{in}}^i$ and ${q_{in}}^j$ in the same side, the formula $\nabla f({q_{in}}^i) \cdot \nabla f({q_{in}}^j) > 0$ holds true. So, we can classify whether two points are in the same or the opposite side using dot product of gradients, $cls(q_i, q_j) = \nabla f(q_i) \cdot \nabla f(q_j)$. Based on that, we divide the space into 3D grids (e.g. $256^3$), and perform gradient discrimination on the 8 vertices $v_i$ in each cell grid according to $cls(q_i, q_j)$. As shown in Fig. \ref{fig:mcubes}(a), the gradient field separates the vertices into two sets, where we can further adapt the marching cubes algorithm \cite{lorensen1987marching} to create triangles for the grid using the lookup table. The complete surface is generated by grouping triangles of each grid together. To accelerate the surface extraction process and avoid extracting unexpected triangles in the multi-layers structures, we set a threshold $\theta$ to stop surface extraction on grids where $f(g_i)>\theta, i \in [0,7]$.

\noindent \textbf{Mesh refinement.} The initial surface extracted by the gradient-based surface extraction algorithm is only a discrete approximation of the zero iso-surface. To achieve a more detailed mesh, we propose to refine it using the UDF values. As shown in Fig. \ref{fig:mcubes}(b), given the predicted UDF values $UDF_A$ and $UDF_B$ of grid vertices $v_A$ and $v_b$, the mesh vertex $v_M$ can be moved to a finer position where $l_{A\rightarrow M} : l_{M\rightarrow B} = UDF_A : UDF_B$. \md{The $l_{A\rightarrow M}$ and $l_{M\rightarrow B}$ indicate the distance from $v_A$ to $v_M$ and the distance from $v_M$ to $v_B$. In simpler terms, the mesh refinement strategy is to move the mesh vertices on to the zero-level set based on a linear interpolation using the UDF values. }

\noindent {\textbf{Surface re-orientation.}
\md{Another issue is that the global consistency of extracted meshes cannot be guaranteed since the surfaces are not closed. This is a common issue for most existing approaches in reconstructing open surfaces from point clouds, such as NDF \cite{chibane2020neural}, MeshUDF \cite{guillard2021meshudf} and GIFS \cite{ye2022gifs}. To solve this issue, we further propose a novel approach to re-direct the surface orientations with the help of non-zero level sets from the learned CAP-UDF. Our insight comes from the observation that although the open surfaces extracted from the zero-level set of learned CAP-UDF are orientation ambiguous, the surfaces extracted from the non-zero level sets with marching cubes algorithm \cite{lorensen1987marching} are usually closed and with correct normal orientations. For example, given a non-zero level set where UDF = 0.01, we can use marching cubes algorithm to obtain a double-layer closed mesh with a thickness of 0.02. Driven by this observation, we propose to leverage the closed surfaces from the non-zero level set of CAP-UDF as the guidance to re-direct the normal orientations on the ones extracted from the zero level set. We show the overview of surface re-orientation process in Fig. \ref{fig:mesh_reorient}.
}

\begin{figure}[tb]
    \centering
    \includegraphics[width=0.5\textwidth]{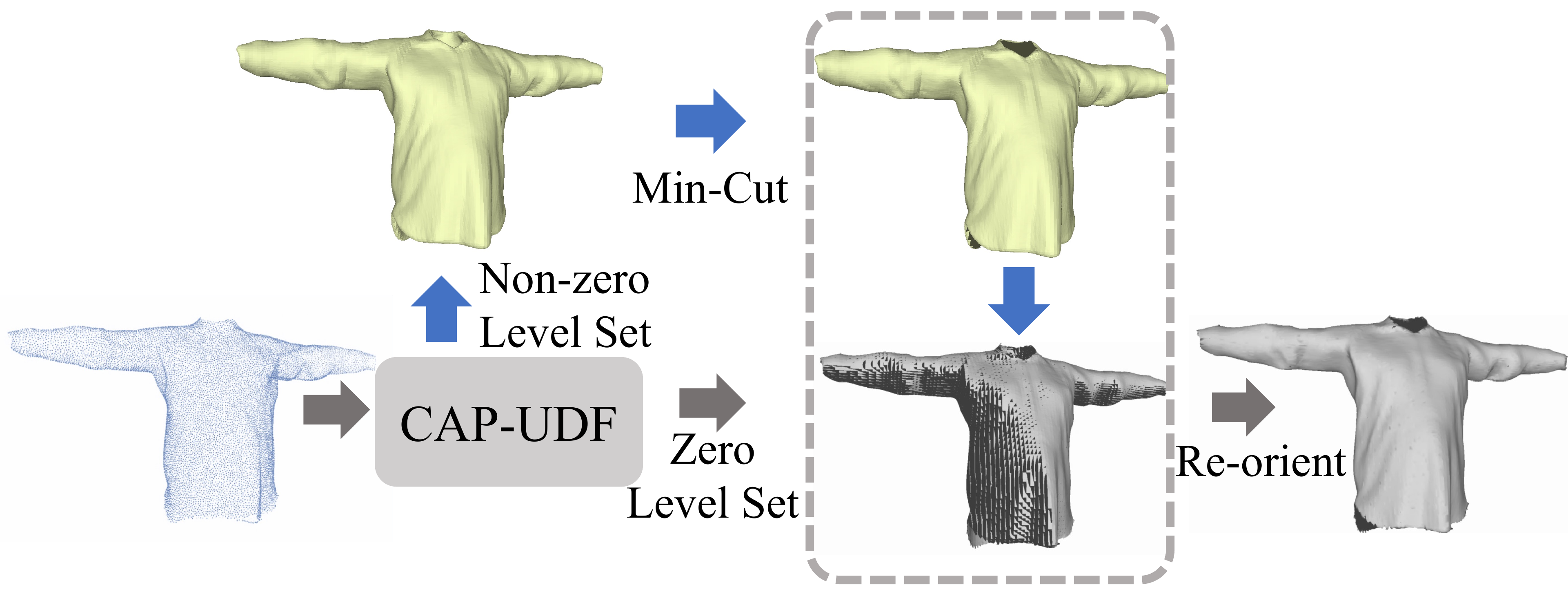}
    \vspace{-0.8cm}

    \caption{\md{Surface re-orientation algorithm for the extracted open surfaces. We extract a double-layer closed mesh from the non-zero level set of learned CAP-UDF with marching cubes \cite{lorensen1987marching}, and cut the mesh with the Min-Cut algorithm \cite{hou2023robust} to only keep the outer layer surfaces, shown as the yellow meshes. We then leverage the outer layer surfaces as a guidance to re-orient the open surfaces extracted from the zero-level set.}}
    \label{fig:mesh_reorient}
\end{figure}

\md{Given an input point cloud $p$, we learn an unsigned distance field from it with the proposed CAP-UDF. The open mesh surfaces $S_{0}$ are extracted from the zero level set with our proposed surface extraction algorithm, as described in Fig. \ref{fig:mcubes} and Sec. \ref{sec:3.4}. The mesh can be leveraged in some downstream tasks for AR, VR, physical simulation, etc. However, the surface orientation may not be correct due to the open structure, leading to negative affect on the downstream rendering and lighting applications. To solve this issue, we simultaneously extract another mesh $S_{\sigma}$ from the non-zero level set of the learned distance field by using the marching cubes algorithm with a non-zero threshold $\sigma$ (e.g. 0.01). The mesh $S_{\sigma}$ is usually a closed surface which has consistent normal orientations. We then cut the double-layer closed mesh $S_{\sigma}$ with the Min-Cut algorithm used in DCUDF \cite{hou2023robust}, and keep the outer layer mesh surfaces $S_{\sigma}^{out}$ as the guidance to re-direct the normals on the surface from the zero level set. 
Specifically, for each vertex $\{v_0^{i}\}_{i=1}^N$ on $S_{0}$, we re-direct its normal $\{n_0^{i}\}_{i=1}^N$ by first searching for its closest vertex $v_{\sigma}’$ on $S_{\sigma}$, and then re-direct $n_0^{i}$ as:}

\begin{equation}
    n_0^{i} = \begin{cases}
        n_0^{i}, &\text{if  \,} n_0^{i} \times n_{\sigma}’ > 0 \\
        - n_0^{i}, &\text{if  \,} n_0^{i} \times n_{\sigma}’ < 0 \\
    \end{cases}
\end{equation}

\md{
Note that we do not directly adopt the outer layer surfaces as the final result since they are from a non-zero level set of the learned UDF, which do not represent the actual surfaces.
}

\subsection{Extension to Unsupervised Point Normal Estimation}
\label{sec:3.5}
\begin{figure}[h]
    \centering
    \includegraphics[width=0.45\textwidth]{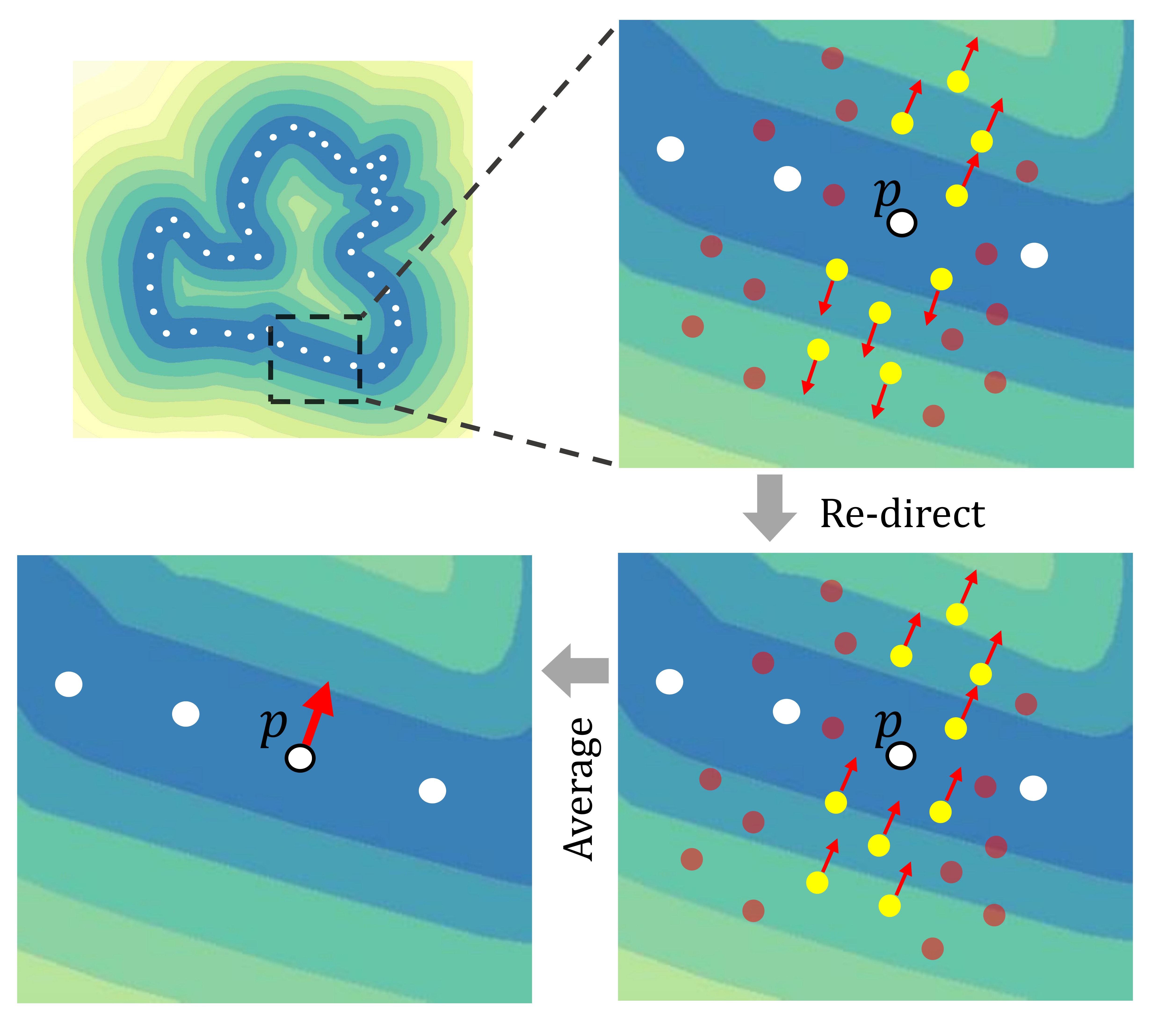}
    \caption{Illustration of unsupervised point normal estimation. The white points indicates the raw point cloud $P$. To estimate the normal for point $p$, we first sample a set of queries near $P$ and search for the subset of queries $Q'$ whose nearest point on $P$ is $p$ (indicated as the yellow points). We then compute gradients at $Q'$ and re-direct them. The normal highlighted as the red arrow at $p$ is extracted by averaging the re-directed gradients.  }
    \label{fig:normalmethod}
\end{figure}

\noindent Normal estimation for unstructured point clouds is to predict a normal for each point in a point cloud. Previous methods \cite{ben2020deepfit, li2022neaf, li2022hsurf} achieve encouraging results by using large-scale annotated ground truth normal labels for supervised training, which are hard to collect. We show that our method can also be extended to estimate the normals for point clouds in an unsupervised way. 
\md{Following most of the previous works \cite{ben2020deepfit, li2022neaf, ben2019nesti} on point normal estimation, we focus on estimating unoriented point cloud normals with neural networks. This means that we only care about if the predicted normals are collinear with the GT point normals, rather than the same direction. The globally consistent normal orientation can be further achieve by applying some off-the-shelf normal orientation methods, e.g. ODP \cite{metzer2021orienting} as a post-processing procedure on the estimated normals.}

A simple implementation is to take the gradient $\nabla f(p)$ at $p$ of the raw point cloud $P$ as the predicted normal. However, it is not differentiable at the zero-level set of UDF, which leads to an unreliable gradient at the surface. We resolve this problem by fusing the gradients of nearby queries. Specifically, given a well-optimized unsigned distance field which is learned from the raw point cloud, we can estimate the normal for each point with a set of query points around it. To extract the normal for a point $p$ in $P$, we sample a set of queries $Q'={q_i'}, {i\in [1,K]}$, whose nearest point on $P$ is $p$. Since we learn a continuous unsigned distance field around $p$, the normal $n$ of $p$ can be approximated as the fusion of the gradients $g_i' = \nabla f(q_i'), {i\in [1,K]}$ in $Q'$. A direct implementation is to calculate the average of $g_i'$, formulated as:

\begin{equation}
    \label{eq:normal}
    n = \frac{1}{K} \sum_{q' \in Q'}{\nabla f(q')}.
\end{equation}

However, the simple averaging described above will lead to a large error since $q_i', i\in [1,K]$ may lie in different side of the surface, which leads to the uncertain sign of $g_i'$. To solve this issue, we propose to normalize the signs of $g_i', i\in [1,K]$ before averaging them together. Specifically, we randomly choose one query $q_v'$ in $Q'$ and take the gradient $g_v'$ as the reference to re-direct the other gradients according to the dot product between $q_v'$ and $q_i', i\in [1,K]$ as below:

\begin{equation}
    \label{eq:normal2}\md{
    \hat{g_i'} = \phi(q{_v'}{^T} \cdot q_i') g_i',}
\end{equation}
where $\phi$ is the sign function with an output in $\{-1, 1\}$.

\begin{figure}[tb]
    \centering
    \includegraphics[width=0.48\textwidth]{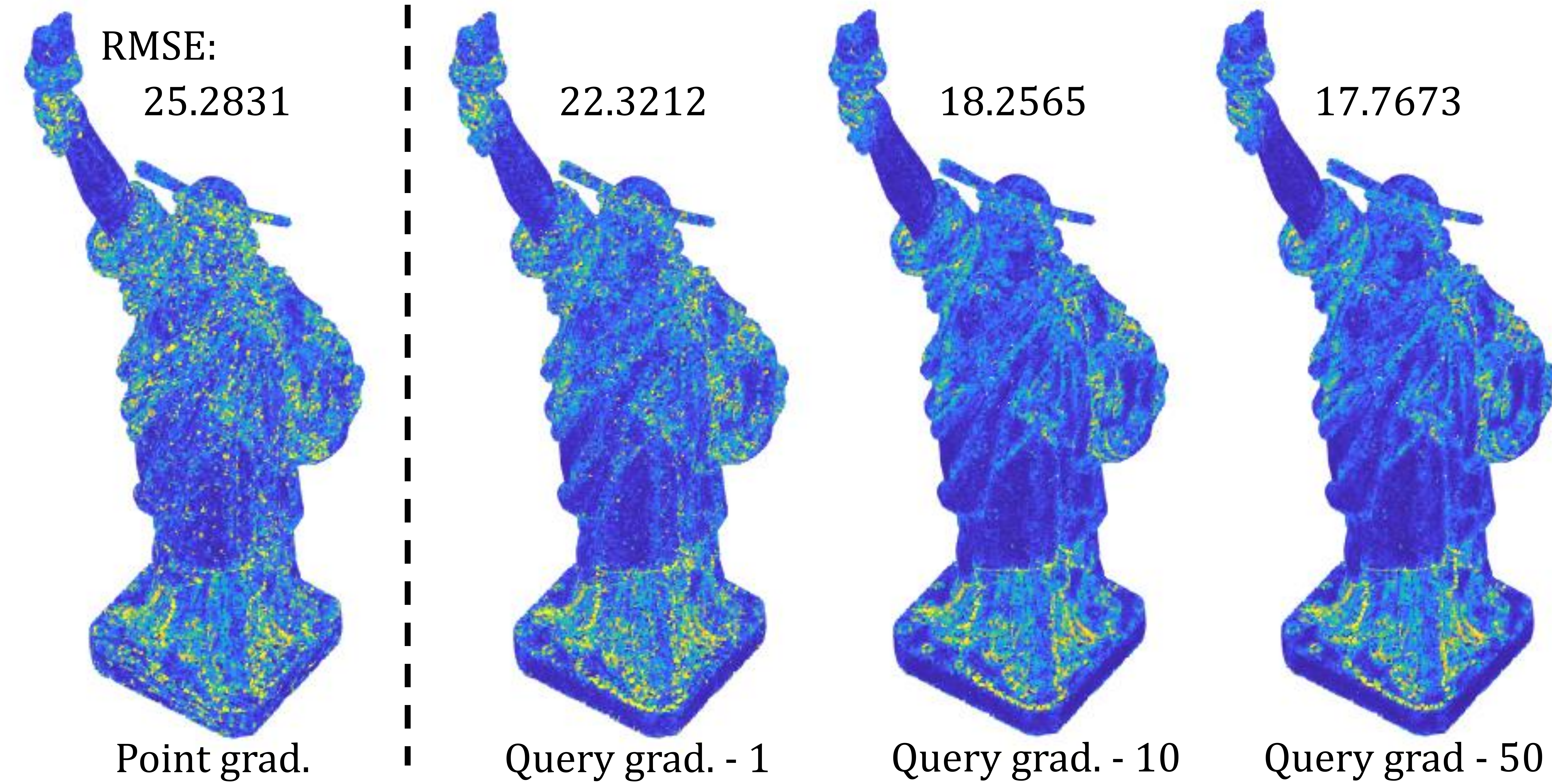}
    \caption{Errors of normal estimation with different implementations. The ``Point grad.'' means taking the gradient at the raw point cloud as normals and ``Query grad. - $K$'' means taking the fusion of gradients at $K$ selected queries as normals. The bluer the color, the smaller the error. }
    \label{fig:normalfig}
\end{figure}

We illustrate the normal estimation algorithm in Fig. \ref{fig:normalmethod}. To demonstrate the effectiveness of our proposed query gradient fusion strategy, we further provide the visual comparison with the baseline as shown in Fig. \ref{fig:normalfig}. The query sampling strategy is described in the implementation details of Sec. \ref{sec:4}. For more experimental comparisons and ablation studies, please refer to Sec. \ref{section:normal} and Sec. \ref{section:normalablation}.

\begin{figure*}[!tb]
  \centering
  \includegraphics[width=2\columnwidth]{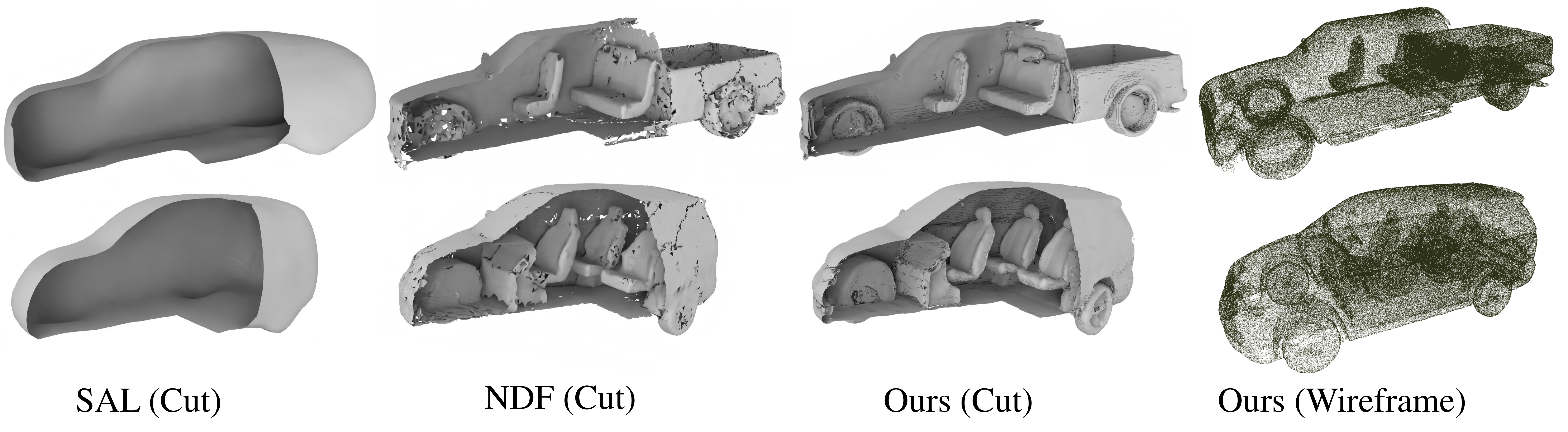}
  \caption{Visual comparisons of surface reconstruction on ShapeNet cars.}
  \label{fig:vis_car}
\end{figure*}

\section{Experiments}
\label{sec:4}
We evaluate our performance in surface reconstruction from raw point clouds or depth maps, and extend our method to unsupervised point normal estimation. We first demonstrate the ability of our method to reconstruct general shapes with open and multi-layer surfaces in Sec. \ref{section:sec4.1}. Next, we apply our method to reconstruct surfaces for real scanned raw data including 3D objects in Sec. \ref{section:sec4.2} and complex scenes in Sec.  \ref{section:sec4.3}. We then extend our method to reconstruct surfaces from the depth maps in Sec. \ref{section:depth} and the point normal estimation task in Sec. \ref{section:normal}. Ablation studies are shown in Sec. \ref{section:sec4.4}. Finally, we provide the efficiency analysis and more visualizations in Sec. \ref{section:efficiency} and Sec. \ref{section:more}.

\noindent\textbf{Implementation details.} 
To learn UDFs for raw point clouds $P$, we adopt a neural network similar to OccNet \cite{mescheder2019occupancy} to predict the unsigned distance given 3D queries as input. Our network contains 8 layers of MLP where each layer has 256 nodes. 
We adopt a skip connection in the fourth layer as employed in DeepSDF \cite{park2019deepsdf} and ReLU activation functions in the last two layers of MLP. To make sure the network learns an unsigned distance, we further adopt a non-linear projection $g(x)=|x|$ before the final output.

Similar to Neural-Pull and SAL, given the single point cloud $P$ as input, we do not leverage any condition and overfit the network to approximate the surface of $P$ by minimizing the loss of Eq. (\ref{eq:gcloss}). Therefore, we do not need to train our network on large scale training dataset in contrast to previous methods \cite{chibane2020neural, ye2022gifs, jiang2020local}. In addition, we use the same strategy as Neural-Pull to sample 60 queries around each point $p_i$ on $P$ as training data. A Gaussian function $\mathcal{N}(\mu, \sigma^2)$ is adopt to calculate the sampling probability where $\mu=p_i$ and $\sigma$ is the distance between $p_i$ and its 50-th nearest points on $P$. The queries used to estimate the normals are sampled in the same way as above.
For sampling auxiliary points in the low confidence region, the standard deviation is set to $1.1\sigma$. And the size of $K$ for estimating normals in Eq. \ref{eq:normal} is set to 50.

During training, we employ the Adam optimizer with an initial learning rate of 0.001 and a cosine learning rate schedule with 1k warn-up iterations. We start the training of next stage after the previous one converges. In experiment, we found the iteration of 40k, 60k and 70k suitable for the change of stages. Since the 3rd and 4th stages bring little advance as shown in Tab. \ref{tab:ablation_step} of the ablation studies, we specify the number of stages as two in practical, so the network has been trained in total 60k iterations. For the surface reconstruction of real scanned complex scenes, we increase the iterations to 300k for a better convergence.

\begin{figure}[!t]
    \centering
    \includegraphics[width=\columnwidth]{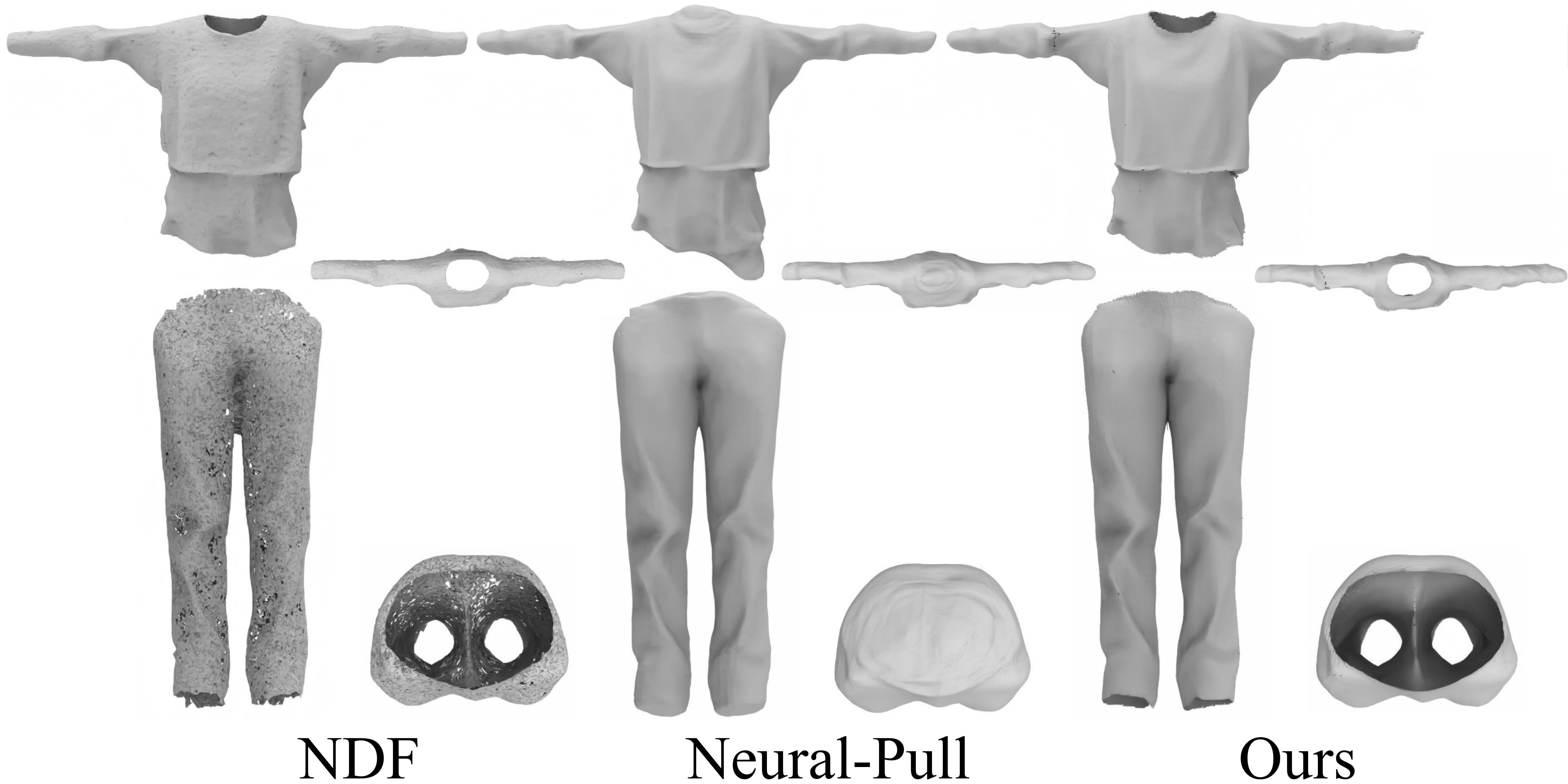}
    \caption{Visual comparisons of surface reconstruction on MGN dataset.}
    \label{fig:mgd}
\end{figure}

\begin{table}
    \caption{Surface reconstruction for point cloud on ShapeNet cars dataset (Chamfer-L2$\times 10^4$).}
    \centering
\resizebox{\linewidth}{!}{
\begin{tabular}{l|cc|cc}
\toprule
\multirow{1}*{} & \multicolumn{2}{c}{Chamfer-L2} & \multicolumn{2}{c}{F-Score}\\

Method  &Mean  &Median & $F1^{0.005}$ & $F1^{0.01}$\\ 
\midrule
Input & 0.363 & 0.355 &48.50 &88.34 \\
Watertight GT & 2.628 & 2.293 & 68.82 & 81.60\\
GT & 0.076 & 0.074 & 95.70 & 99.99\\
NDF$_{BPA}$ \cite{chibane2020neural} & 0.202 & 0.193 &77.40 & 97.97\\ 
NDF$_{gradRA}$ & 0.160 & 0.152 &82.87 & 99.35\\
NDF$_{PC}$ & 0.126 & 0.120 & 88.09 & 99.54\\
GIFS \cite{ye2022gifs} & 0.128 & 0.123 & 88.05  & 99.31 \\
\midrule
Ours$_{BPA}$ & 0.141 & 0.138 & 84.84 & 99.33 \\
Ours$_{gradRA}$ & \textbf{0.119} & \textbf{0.114} & \textbf{88.55} & \textbf{99.82} \\
Ours$_{PC}$ & \textbf{0.110} & \textbf{0.106} & \textbf{90.06} & \textbf{99.87} \\
\bottomrule
\end{tabular}}

    \label{table:cars}
\end{table}

\subsection{Surface Reconstruction for Synthetic Shapes}
\label{section:sec4.1}
\textbf{Dataset and metrics.} 
For the experiments on synthetic shapes, we follow NDF \cite{chibane2020neural} to choose the ``Car" category of the ShapeNet dataset which contains the greatest amount of multi-layer shapes and non-closed shapes. And 10k points is sampled from the surface of each shape as the input. Besides, we employ the MGN dataset \cite{bhatnagar2019multi} to show the advantage of our method in open surfaces. To measure the reconstruction quality, we follow GIFS \cite{ye2022gifs} to sample 100k points from the reconstructed surfaces and adopt the Chamfer distance ($\times10^4$), Normal Consistency (NC) \cite{mescheder2019occupancy} and F-Score with a threshold of 0.005/0.01 as evaluation metrics.

\noindent\textbf{Comparison.}
We compare our method with the state-of-the-art works NDF \cite{chibane2020neural} and GIFS \cite{ye2022gifs}. We quantitatively evaluate our method with NDF and GIFS in Tab. \ref{table:cars}. We also report the results of points sampled from the watertight ground truth (watertight GT in table) as the upper bound of the traditional SDF-based or Occupancy-based implicit functions. To show the superior limit of this dataset, we sample two different sets of points from the ground truth mesh and report their results (GT in table). For a comprehensive comparison with NDF, we transfer our gradient-based reconstruction algorithm to extract surfaces from the learned distance field of NDF, and report three metrics of NDF and our method including generated point cloud ($*_{PC}$), mesh generated using BPA ($*_{BPA}$) and mesh generated using our gradient-based reconstruction algorithm ($*_{gradRA}$). As shown in Tab. \ref{table:cars}, we achieve the best results in terms of all the metrics. Moreover, our gradient-based reconstruction algorithm shows great generality in transferring to the learned gradient field of other method (e.g. NDF) by achieving significant improvement over the traditional method (BPA). We also provide the results of surface reconstruction on MGN \cite{bhatnagar2019multi} dataset as shown in Tab. \ref{table:mgd}, where we significantly outperform other methods.

We further present a visual comparison with SAL and NDF in Fig. \ref{fig:vis_car}. Previous methods (e.g. SAL) take SDF as output and are therefore limited to single-layer shapes where the inner-structure is lost. NDF learns UDFs and is able to represent general shapes, but it outputs a dense point cloud and requires BPA to generate meshes, which leads to an uneven surface. On the contrary, we can extract surfaces directly from the learned UDFs, which are continuous surfaces with high fidelity. 
We also provide a visual comparison with Neural-Pull in MGN dataset as shown in Fig. \ref{fig:mgd}, where we accurately reconstruct the open surfaces but Neural-Pull fails to reveal the original geometry.

\subsection{Surface Reconstruction for Real Scans}
\label{section:sec4.2}
\textbf{Dataset and metrics.}
For surface reconstruction of real point cloud scans, we follow SAP to evaluate our methods under the Surface Reconstruction Benchmarks (SRB) \cite{williams2019deep}. We use Chamfer distance and F-Score with a threshold of 1\% for evaluation. Note that the ground truth is dense point clouds.

\begin{figure*}[!t]
  \centering
  \includegraphics[width=2\columnwidth]{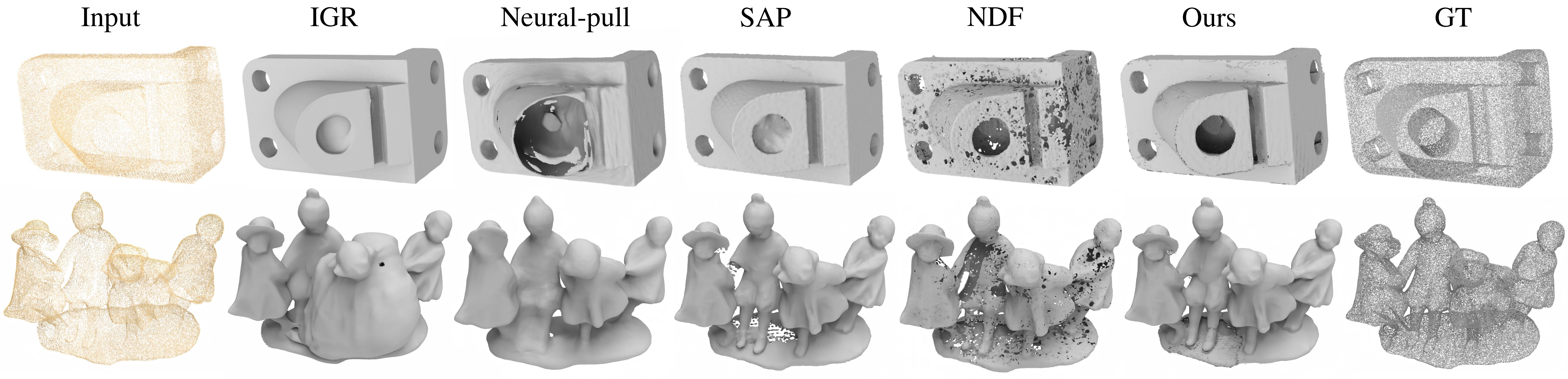}
  \caption{Visual comparisons of surface reconstruction on the SRB dataset.}
  \label{fig:vis_spb}
\end{figure*}

\begin{table}[tb]
    \caption{Surface reconstruction for point cloud on MGN dataset.} 
    \centering
    \setlength{\tabcolsep}{2mm}
    \resizebox{\linewidth}{!}{
    \begin{tabular}{l|c|c|c}
    \toprule
    
    Method  &Chamfer-L2  &F-Score$^{0.01}$ & NC\\ 
    \midrule
    Neural-Pull \cite{ma2021neural} & 4.447 & 94.49 & 91.83\\
    NDF \cite{chibane2020neural} & 0.658 & 76.11 & 92.84\\
    Ours & \textbf{0.117} & \textbf{99.68} & \textbf{97.80}\\
    \bottomrule
    \end{tabular}}

    \label{table:mgd}
\end{table}

\noindent\textbf{Comparison.} We compare our method with state-of-the-art classic and data-driven surface reconstruction methods in the real scanned SRB dataset, including IGR \cite{gropp2020implicit}, Point2Mesh \cite{hanocka2020point2mesh}, Screened Poisson Surface Reconstruction (SPSR) \cite{kazhdan2013screened}, Shape As Points (SAP) \cite{peng2021shape}, Neural-Pull \cite{ma2021neural} and NDF \cite{chibane2020neural}. The numerical comparison is shown in Tab. \ref{table:srb}, where we achieve the best accuracy. The visual comparisons in Fig. \ref{fig:vis_spb} demonstrate that our method is able to reconstruct a continuous surface with local geometry consistence while other methods struggle to reveal the geometry details. For example, IGR, Neural-Pull and SAP mistakenly mended or failed to reconstruct the hole of the anchor while our method is able to keep the correct geometry.

\begin{table}[h]
    \caption{Surface reconstruction for point cloud on SRB dataset.}
    \centering
    \setlength{\tabcolsep}{4mm}
\resizebox{0.95\linewidth}{!}{
    \begin{tabular}{l|c|c}
    \toprule
    
    Method  &Chamfer-L1  &F-Score\\ 
    \midrule
    
    IGR \cite{gropp2020implicit} & 0.178 & 75.5 \\
    Point2Mesh \cite{hanocka2020point2mesh} & 0.116 & 64.8\\
    SPSR \cite{kazhdan2013screened} & 0.232 & 73.5\\
    SAP \cite{peng2021shape} & 0.076 & 83.0\\
    Neural-Pull \cite{ma2021neural} & 0.106 & 79.7\\
    NDF$_{PC}$ \cite{chibane2020neural} & 0.185 & 72.2\\
    NDF$_{mesh}$ & 0.238 & 68.6\\
    \midrule
    Ours$_{PC}$ & \textbf{0.068} & \textbf{90.4}\\
    Ours$_{mesh}$ & \textbf{0.073} & \textbf{84.5}\\
    \bottomrule

    \end{tabular}}

    \label{table:srb}
\end{table}

\subsection{Surface Reconstruction for Scenes}
\label{section:sec4.3}
\textbf{Dataset and metrics.}
To further demonstrate the advantage of our method in surface reconstruction of real scene scans, we follow OnSurf \cite{On-SurfacePriors} to conduct experiments under the 3D Scene dataset \cite{zhou2013dense}. Note that the 3D Scene dataset is a challenging real-world dataset with complex topology and noisy open surfaces. We uniformly sample 100, 500 and 1000 points per $m^2$ at the original scale of scenes as the input and follow OnSurf to sample 1M points on both the reconstructed and the ground truth surfaces for evaluation. We leverage L1 and L2 Chamfer distance to evaluate the reconstruction quality.

\noindent\textbf{Comparison.} 
We compare our method with the state-of-the-arts scene reconstruction methods ConvONet \cite{peng2020convolutional}, LIG \cite{jiang2020local}, DeepLS \cite{chabra2020deep}, OnSurf \cite{On-SurfacePriors} and NDF \cite{chibane2020neural}. The numerical comparisons in Tab. \ref{table:scenes} show that our method significantly outperform the other methods under different point densities. The visual comparisons in Fig. \ref{fig:vis_scene} further shows that our reconstructions present more geometry details in complex real scene scans. Note that all the other methods have been trained in a large scale dataset, from which they gain additional prior information. On the contrary, our method does not leverage any additional priors or large scale training datasets, and learns to reconstruct surfaces directly from the raw point cloud, but still yields a non-trivial performance.

\begin{table}[h]
			\caption{Surface Reconstruction under 3D Scene, L2CD$\times$1000.}
\centering
\vspace{-0.2cm}
\resizebox{\linewidth}{!}{
            \begin{tabular}{l|cc|cc|cc}
            \hline
            \multirow{2}*{Method} & \multicolumn{2}{c}{100/$m^2$} & \multicolumn{2}{c}{500/$m^2$} & \multicolumn{2}{c}{1000/$m^2$}\\
            \cline{2-7}
            ~ & L2CD & L1CD & L2CD & L1CD & L2CD & L1CD\\
            \hline
            ConvONet \cite{peng2020convolutional} & 7.859 & 0.043 & 13.192 & 0.052 & 14.097 & 0.052 \\
            LIG \cite{gropp2020implicit} & 6.265 & 0.049 & 5.633 & 0.048 & 6.190 & 0.048  \\
            DeepLS \cite{chabra2020deep} & 3.029 & 0.044 & 6.794 & 0.050 & 1.607 & 0.025 \\
            NDF$_{PC}$ \cite{chibane2020neural} & 0.409 & 0.012  & 0.377 & 0.014 & 0.561 & 0.017 \\
            NDF$_{mesh}$ & 0.452 & 0.014 & 0.475 & 0.016 & 0.872 & 0.022 \\
            OnSurf \cite{On-SurfacePriors} & 1.154 & 0.021 & 0.862 & 0.020 & 0.706 & 0.020 \\
            \hline
            Ours$_{PC}$ & \textbf{0.144} & \textbf{0.010} & \textbf{0.078} & \textbf{0.009}  & \textbf{0.072} & \textbf{0.010} \\
            Ours$_{mesh}$ & \textbf{0.187} & \textbf{0.011} & \textbf{0.122} & \textbf{0.010} & \textbf{0.121} & \textbf{0.009} \\
            \hline
            
            \end{tabular}}

			\label{table:scenes}
\end{table}

\begin{figure*}[!t]
  \centering
  \includegraphics[width=2\columnwidth]{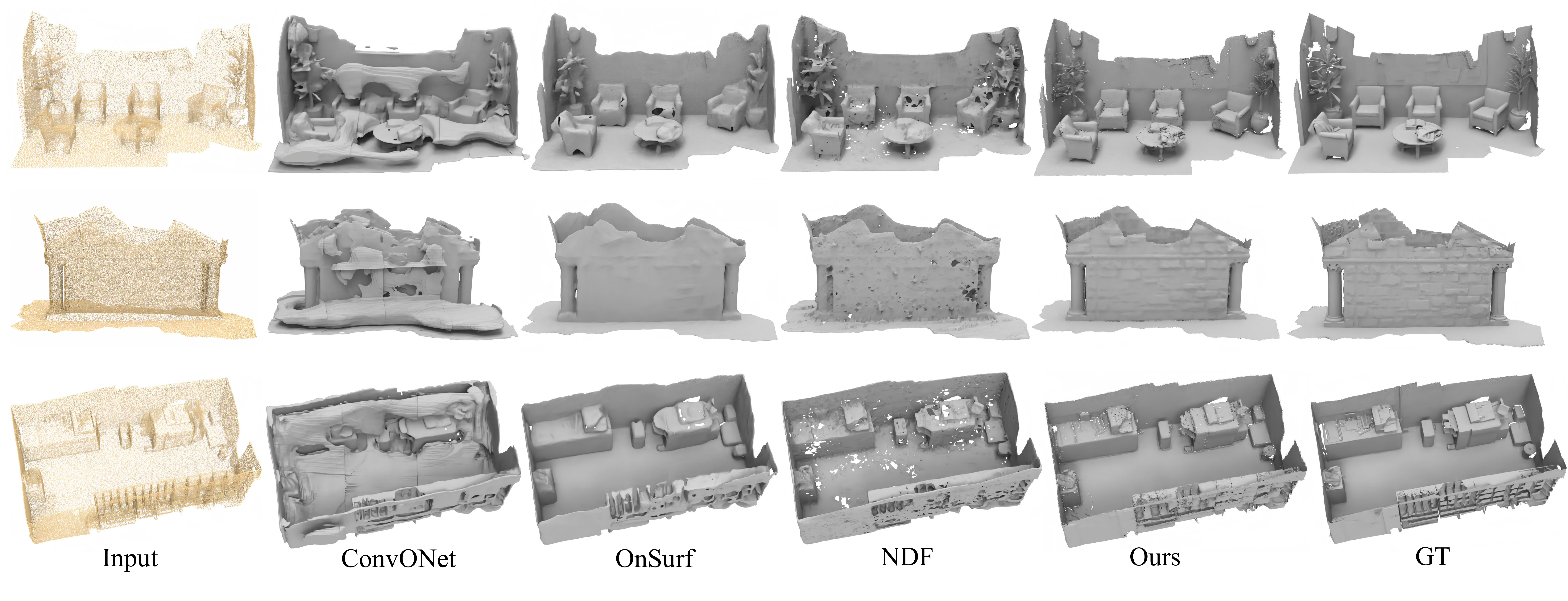}
  \caption{Visual comparison with different methods under 3D Scene. Inputs contains 1k points/$m^2$.}
  \label{fig:vis_scene}
  \vspace{-0.5cm}
\end{figure*}

\subsection{Surface Reconstruction from Depth Maps}
\label{section:depth}

\textbf{Dataset and metrics.} For surface reconstruction from depth maps, we follow NeuralRGB-D \cite{azinovic2022neural} to evaluate our method under the 10 synthetic scenes dataset. To measure the reconstruction quality, we follow NeuralRGB-D and Go-Surf \cite{wang2022go} to sample point clouds at a density of 1 point per $cm^2$ and adopt the Chamfer distance, Normal Consistency(NC) and F-Score with a threshold of 5cm as evaluation metrics.

\noindent\textbf{Comparison.} We compare our method with several state-of-the-art traditional or learning-based methods: BundleFusion \cite{dai2017bundlefusion}, RoutedFusion \cite{weder2020routedfusion}, COLMAP \cite{schonberger2016structure} with Possion Reconstruction (PR) \cite{kazhdan2013screened}, NeRF \cite{mildenhall2020nerf} with additional depth supervision, Convolutional Occupancy Networks (COcc) \cite{peng2020convolutional}, SIREN \cite{sitzmann2020implicit}, Neural-Pull \cite{ma2021neural}, NeuralRGB-D \cite{azinovic2022neural} and Go-Surf \cite{wang2022go}. For a comprehensive comparison, we conduct the scene fusion experiments under the ``Clean'' setting and the ``Noise'' setting. For the ``Clean'' setting where the ground truth camera poses and clean depth maps are known, we simply back-project the depth maps into world space with the camera poses and fuse them together to achieve a global point cloud, and leverage our model to reconstruct surfaces. To further evaluate the ability of our method to handle noises in the real world situation, we follow NeuralRGB-D to apply noise and artifacts to the depth maps to simulate a real depth sensor, which is the ``Noise'' setting. And the ground truth camera poses are also unavailable in the ``Noise'' setting, where we adopt the estimated poses from Go-Surf as the initialization.

The numerical comparison in Tab. \ref{table:nrgbd} shows that our method significantly outperform the other depth-only methods. We also achieve comparable performance with the state-of-the-art RGBD-based methods NeuralRGB-D and Go-Surf, where both depth maps and colored images are required as inputs. We further present a visual comparison under the ``Clean'' setting with COcc, Neural-Pull and NeuralRGB-D in Fig. \ref{fig:vis_nrgbd}. Previous methods takes occupancy (COcc) or SDF (Neural-Pull and NeuralRGB-D) as the scene representation and is limited to the closed geometries, while our method can reconstruct surfaces with arbitrary architecture (e.g. open windows and thin table legs) and also reveal the geometry details. 
\begin{figure*}[h]
  \centering
  \includegraphics[width=2\columnwidth]{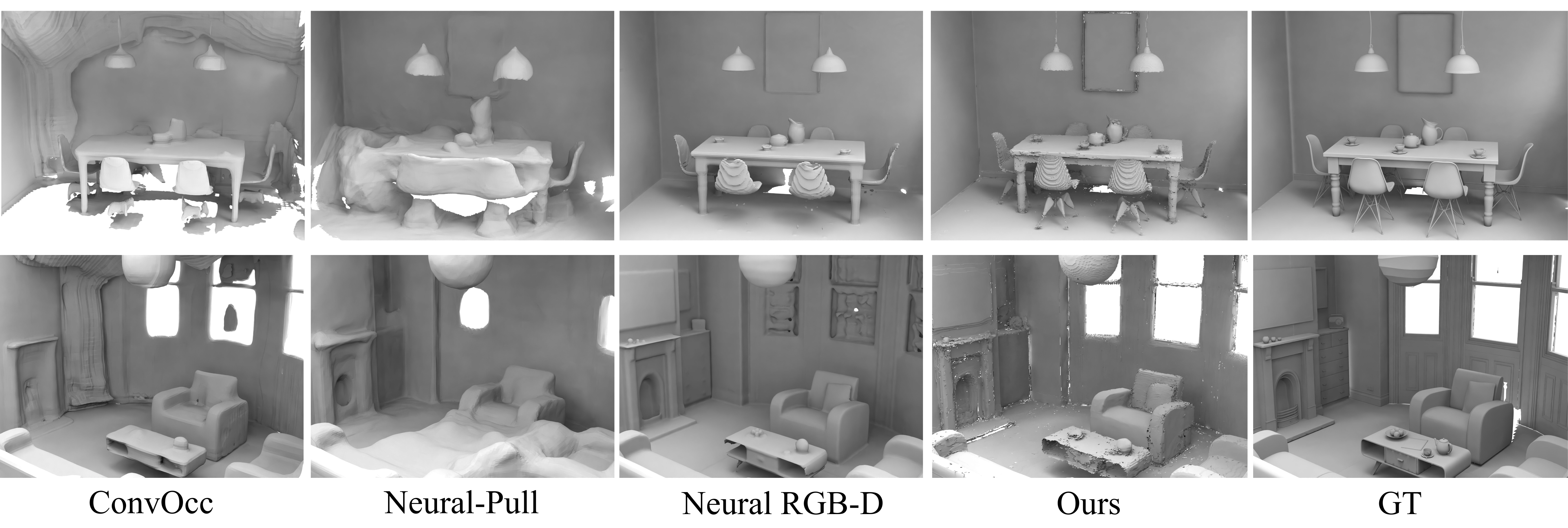}
  \caption{Visual comparison of surface reconstruction from depth maps on the 10 synthetic scenes dataset.}
  \label{fig:vis_nrgbd}
\end{figure*}

\begin{table}
\caption{Surface reconstruction for depth maps on 10 synthetic scenes dataset.}
\begin{center}
\setlength{\tabcolsep}{1.5mm}
\resizebox{\linewidth}{!}{
\begin{tabular}{c|l|c|c|c|c}
\hline\noalign{\smallskip}
~ & Method & Input & L1CD & NC & F-Score\\
\noalign{\smallskip}
\hline
\noalign{\smallskip}
\multirow{8}*{\rotatebox{90}{Noise}} 
& BundleFusion \cite{dai2017bundlefusion} & RGB-D & 0.062 & 0.892 & 0.805\\
~ & COLMAP\cite{schonberger2016structure}+PR \cite{kazhdan2013screened} & RGB-D & 0.057 & 0.901 & 0.839\\
~ & NeRF+Depth \cite{mildenhall2020nerf} & RGB-D &0.065 & 0.768 & 0.782\\
~ & NeuralRGB-D \cite{azinovic2022neural} & RGB-D &0.044 & 0.918 & 0.924\\
~ & RoutedFusion \cite{weder2020routedfusion} & Depth & 0.057 & 0.864 & 0.838\\
~ & COcc \cite{peng2020convolutional} & Depth & 0.077 & 0.849 & 0.643\\
~ & SIREN \cite{sitzmann2020implicit} & Depth & 0.060 & \textbf{0.893} & 0.816\\
~ & Ours & Depth & \textbf{0.052} & 0.880 & \textbf{0.854}\\
\noalign{\smallskip}
\hline
\noalign{\smallskip}
\multirow{5}*{\rotatebox{90}{Clean}} & 
 NeuralRGB-D \cite{azinovic2022neural} & RGB-D & 0.0134 & 0.943 & 0.985\\
~ & Go-Surf \cite{wang2022go} & RGB-D & 0.0102 & 0.942 & 0.983\\
~ & COcc \cite{peng2020convolutional} & Depth & 0.0943 & 0.860 & 0.693\\
~ & Neural-Pull \cite{ma2021neural} & Depth & 0.0821 & 0.884 & 0.761\\
~ & Ours & Depth & \textbf{0.0081} & \textbf{0.934} & \textbf{0.987} \\
\noalign{\smallskip}

\hline

\end{tabular}}
\end{center}

\label{table:nrgbd}
\end{table}

\subsection{Point Normal Estimation}
\label{section:normal}
\textbf{Dataset and metrics.} For the point cloud normal estimation task, we adopt the widely-used benchmark PCPNet \cite{guerrero2018pcpnet} dataset to evaluate our method. PCPNet samples 100k points on the mesh of each shape to obtain a point cloud. In addition to the ``Clean'' data which is sampled uniformly, two additional data with varying density settings (Stripes and Gradients) are added to evaluate the ability of different methods in handling irregular data. We aim at learning unsigned distance fields for unsupervised point cloud normal estimation, thus only the test set of PCPNet dataset is used. We adopt the angle root mean square error (RMSE) between the predicted normals and the ground truth normals as the metric to evaluate the performance of our method.  Following previous works \cite{ben2020deepfit, li2022neaf}, we compute the final result with a 5000 points subset for each shape.

\noindent\textbf{Comparison.} 
We conduct a comparison with traditional normal estimation methods including PCA \cite{abdi2010principal}, Jets \cite{cazals2005estimating} and the state-of-the-art learning-based methods PCPNet \cite{guerrero2018pcpnet}, HoughCNN \cite{boulch2016deep}, Nesti-Net \cite{ben2019nesti}, Iter-Net \cite{lenssen2020deep} and DeepFit \cite{ben2020deepfit}. Note that all the previous learning-based methods are designed in a supervised manner where a large scale dataset is required for training. On the contrary, we resolve the problem from a new perspective, where the normals can be extracted directly from the learned unsigned distance fields in an unsupervised manner. The quantitative comparison in Tab. \ref{table:normal} shows that our method significantly outperform the traditional methods which also do not require ground truth normals for supervision. Meanwhile, our method also achieves a comparable or even better performance than the state-of-the-art supervised normal estimation methods which demonstrates that the expensive annotated normal labels are not necessary for point normal estimation task.

We further provide a visual comparison with the unsupervised methods PCA, Jets and the supervised methods PCPNet, Nesti-Net and DeepFit in Fig. \ref{fig:pcp}. The color of the shape indicates errors, where the closer to yellow the larger the error and the closer to blue the smaller the error. As shown, our estimation results are more accurate and detailed compared to other methods, especially on the complex geometries such as sharp edges and corners.

\begin{figure}[h]
  \centering
  \includegraphics[width=\columnwidth]{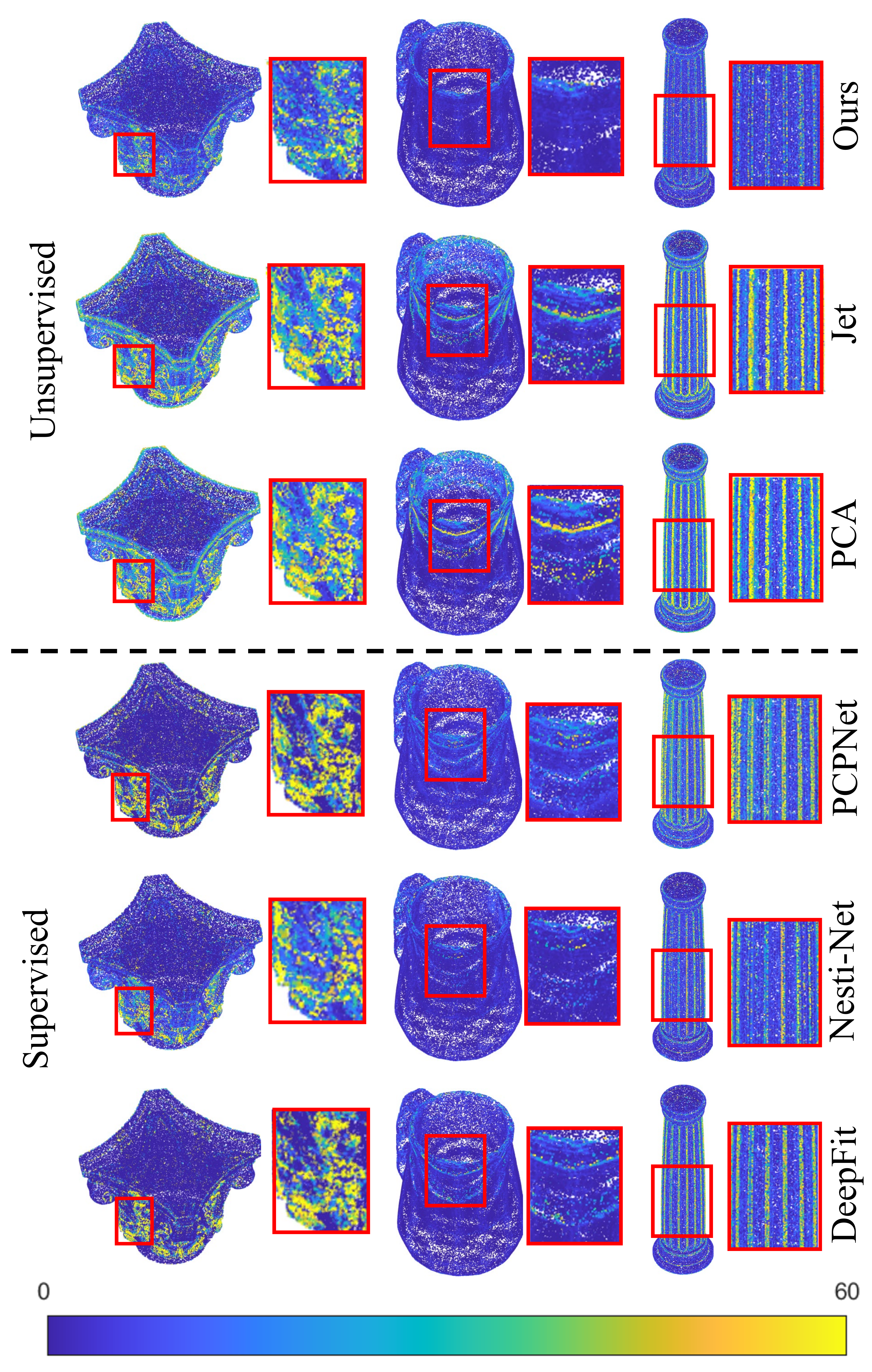}
  \caption{Visual comparison on the error maps of point normal estimation under the PCPNet dataset.}
  \label{fig:pcp}
\end{figure}

\begin{table}
\caption{RMSE of point normal estimation on the PCPNet dataset 
(lower is better).}
\begin{center}
\setlength{\tabcolsep}{1mm}
\resizebox{\linewidth}{!}{
\begin{tabular}{c|l|ccc|c}
\hline\noalign{\smallskip}
~ & Method & Clean & Strip & Gradient & average\\
\noalign{\smallskip}
\hline
\noalign{\smallskip}
\multirow{5}*{Supervised} 
& PCPNet \cite{guerrero2018pcpnet} & 9.66 & 11.47 & 13.42 & 11.61\\
~ & Hough \cite{boulch2016deep} & 10.23 & 12.47 & 11.02 & 11.24\\
~ & Nesti-Net \cite{ben2019nesti} & 6.99 & 8.47 & 9.00 & 8.15\\
~ & IterNet \cite{lenssen2020deep} & 6.72 & 7.73 & 7.51 & 7.32\\
~ & DeepFit \cite{ben2020deepfit} & 6.51 & 7.92 & 7.31 & 7.25\\
\noalign{\smallskip}
\hline
\noalign{\smallskip}
\multirow{3}*{Unsupervised} & 
 Jet \cite{cazals2005estimating} & 12.23 & 13.39 & 13.13 & 12.92\\
~ & PCA \cite{abdi2010principal} & 12.29 & 13.66 & 12.81 & 12.92\\
~ & Ours & \textbf{6.33} & \textbf{7.26} & \textbf{7.11} & \textbf{6.90} \\
\noalign{\smallskip}

\hline

\end{tabular}}
\end{center}

\label{table:normal}
\end{table}

\noindent{\textbf{Oriented Normal Comparison.} 
\md{We further conduct experiments on evaluating the oriented normals, which are achieved by applying the off-the-shelf normal orientation method ODP \cite{metzer2021orienting} as a post-processing procedure on the unoriented normals estimated by different methods. The numerical comparison is shown in Tab. \ref{table:orient_normal}, where CAP-UDF still achieves better oriented normal estimation results compared to the unsupervised and even some supervised methods. Note that PCPNet \cite{guerrero2018pcpnet} can be trained to estimate oriented normals directly, where the result is shown as PCPNet*. All other methods estimated unoriented normals from point clouds, followed by ODP to estimate the orientation for each point normal.}

\md{The supervised method PCPNet* which directly estimates oriented normals from point clouds produces much worse results compared to all other methods which estimate unoriented normals with a post-processing procedure for the orientation consistency. The results demonstrate the necessity of decomposing oriented normal estimation task into the two subtasks to estimate the unoriented normals and then produce global consistent orientations. While we focus on the former one for fair comparisons with most of the previous methods.
}

\begin{table}
\caption{\md{RMSE of oriented point normal estimation on the PCPNet dataset 
(lower is better).}}
\vspace{-0.8cm}
\begin{center}
\setlength{\tabcolsep}{1mm}
\resizebox{\linewidth}{!}{
\begin{tabular}{c|l|ccc|c}
\hline\noalign{\smallskip}
~ & Method & Clean & Strip & Gradient & average\\
\noalign{\smallskip}
\hline
\noalign{\smallskip}
\multirow{3}*{Supervised} 
& PCPNet *  \cite{guerrero2018pcpnet} & 33.34 & 37.95 & 35.44 & 35.58\\
\cmidrule(lr){2-6}
~ & Nesti-Net \cite{ben2019nesti} + ODP & 28.87 & 32.07 & 22.27 & 27.73\\
~ & DeepFit \cite{ben2020deepfit} + ODP & 26.60 & 26.74 & \textbf{20.96} & 24.77\\
\noalign{\smallskip}
\hline
\noalign{\smallskip}
\multirow{3}*{Unsupervised} & 
 Jet \cite{cazals2005estimating} + ODP & 28.08 & 23.86 & 23.96 & 25.30\\
~ & PCA \cite{abdi2010principal} + ODP & 28.96 & 28.70 & 23.00 & 28.89\\
~ & Ours + ODP & \textbf{26.47} & \textbf{23.64} & 21.55 & \textbf{23.89} \\
\noalign{\smallskip}

\hline

\end{tabular}}
\end{center}

\label{table:orient_normal}
\end{table}

\subsection{\md{Surface Reconstruction form Corrupted Data}}
\label{sec.4.6}

\md{The real world point clouds are often sparse or corrupted with noises and occlusions. To further apply CAP-UDF to reconstruct the corrupted point clouds, we propose to introduce priors from SoTA point cloud processing approaches. Our idea is to first improve the quality of point clouds, based on which we learn distance fields for reconstruction. To demonstrate the effectiveness of introducing off-the-shelf point cloud processing approaches as priors, we conduct comprehensive experiments on different datasets to evaluate the performance of CAP-UDF in reconstructing corrupted point clouds.  }

\begin{table}
    \caption{\md{Surface reconstruction for occluded point clouds on PCN dataset (Chamfer-L1$\times 10^3$).}}
    \vspace{-0.3 cm}
    \centering
\resizebox{\linewidth}{!}{
\begin{tabular}{l|cc|cc}
\toprule
\multirow{1}*{} & \multicolumn{2}{c}{Chamfer-L1} & \multicolumn{2}{c}{F-Score}\\

Method  &Mean  &Median & $F1^{0.005}$ & $F1^{0.01}$\\ 
\midrule
ConvOcc \cite{peng2020convolutional} & 47.51 & 37.75 & 15.74 & 38.66 \\
POCO \cite{boulch2022poco} & 50.78 & 44.93 & 12.03 & 20.52 \\
\midrule
ConvOcc \cite{peng2020convolutional} + Prior & 11.46 & 10.53 & 27.12 & 66.12 \\
POCO \cite{boulch2022poco} + Prior & 15.52 &13.09 & 29.24 & 61.89 \\
Ours & \textbf{10.42} & \textbf{8.36} & \textbf{35.63} & \textbf{73.27} \\
\bottomrule
\end{tabular}}
    \label{table:completion_pcn}
\end{table}

\subsubsection{\md{Occluded point cloud surface reconstruction}}

\md{
We conduct experiments under the widely-used PCN \cite{yuan2018pcn} dataset for evaluating the reconstruction performances from occluded point clouds. We introduce the pretrained AdaPoinTr \cite{yu2023adapointr} as the point cloud completion prior and reconstruct surfaces from the completed point clouds.}

\md{
The quantitative comparison on reconstructing occluded point clouds is shown in Tab. \ref{table:completion_pcn}. As shown, the SoTA generalized methods (e.g. ConvOcc \cite{peng2020convolutional}, POCO \cite{boulch2022poco}) fail to produce complete reconstruction from the corrupted point clouds. The reason is that although they are trained under large and diverse datasets, they do not learn the completion knowledge, leading to a failure in generalizing to occluded shapes. While we believe that introducing expert point cloud completion methods as the prior is a more effective way to handle corrupted point clouds. As shown in the lower part of Tab. \ref{table:completion_pcn}, the convincing results are achieved by introducing AdaPoinTr as the prior to complete the corrupted point clouds before feeding into the reconstruction frameworks. With the same completion prior, CAP-UDF achieves the best performance compared to the generalized methods ConvOcc and POCO, highlighting the superiority of CAP-UDF in faithfully reconstructing the completed point clouds. Please refer to Sec. 2.1 of the supplementary for visual comparisons on reconstructing occluded point clouds.
}

\begin{table}
    \caption{\md{Surface reconstruction for noisy point clouds on PUNet dataset (Chamfer-L1$\times 10^3$).}}
    \vspace{-0.3 cm}
    \centering
\resizebox{\linewidth}{!}{
\begin{tabular}{l|cc|cc}
\toprule
\multirow{1}*{} & \multicolumn{2}{c}{Chamfer-L1} & \multicolumn{2}{c}{F-Score}\\

Method  &Mean  &Median & $F1^{0.005}$ & $F1^{0.01}$\\ 
\midrule
ConvOcc \cite{peng2020convolutional} & 22.02 & 20.09 & 5.74 & 32.72 \\
POCO \cite{boulch2022poco} & 19.56 & 19.16 & 7.07 & 38.36 \\
\midrule
ConvOcc \cite{peng2020convolutional} + Prior & 16.01 & 13.89 & 13.78 & 58.85 \\
POCO \cite{boulch2022poco} + Prior & 12.38 &11.04 & 23.11 & 74.79 \\
Ours & \textbf{11.36} & \textbf{10.64} & \textbf{25.49} & \textbf{77.87} \\
\bottomrule
\end{tabular}}

    \label{table:denoise_punet}
\end{table}

\subsubsection{\md{Noisy point cloud surface reconstruction}}
\md{
We conduct experiments under the widely-used PUNet \cite{yu2018pu} dataset with 2\% gaussian noises for evaluating the reconstruction performances from noisy point clouds. We introduce the pretrained IterativePFN \cite{de2023iterativepfn} as the point cloud denoising prior and reconstruct surfaces from the denoised point clouds.}

\md{
The quantitative comparison on reconstructing noisy point clouds is shown in Tab. \ref{table:denoise_punet}, where the SoTA generalized results also fails to produce robust reconstruction from the noisy point clouds, demonstrating that training under large datasets in a generalized way still does not generalize well to unseen point noises. By introducing expert point cloud denoising methods as a prior to clean the corrupted point clouds, more accurate reconstructions can be achieved. CAP-UDF demonstrates superior performance in faithfully reconstructing cleaned point clouds when compared to the generalized methods ConvOcc \cite{peng2020convolutional} and POCO \cite{boulch2022poco}, all using the same denoising prior. Please refer to Sec. 2.1 of the supplementary for visual comparisons on reconstructing noisy point clouds.
}

\begin{table}
    \caption{\md{Surface reconstruction for sparse point clouds from PUGAN dataset where each shape only contains 256 points (Chamfer-L1$\times 10^3$).}}
    \vspace{-0.3cm}
    \centering
\resizebox{\linewidth}{!}{
\begin{tabular}{l|cc|cc}
\toprule
\multirow{1}*{} & \multicolumn{2}{c}{Chamfer-L1} & \multicolumn{2}{c}{F-Score}\\

Method  &Mean  &Median & $F1^{0.005}$ & $F1^{0.01}$\\ 
\midrule
ConvOcc \cite{peng2020convolutional} & 56.52 & 39.91 & 11.37 & 37.96 \\
POCO \cite{boulch2022poco} & 61.91 & 57.35 & 1.83 &  5.45 \\
\midrule
ConvOcc \cite{peng2020convolutional} + Prior & 13.66 & 13.60 & 18.93 & 66.25 \\
POCO \cite{boulch2022poco} + Prior & 10.36 & 10.11 & 33.12 & 82.03 \\
Ours & \textbf{9.44} & \textbf{9.24} & \textbf{39.95} & \textbf{86.18} \\
\bottomrule
\end{tabular}}

    \label{table:upsamp_256}
\end{table}

\begin{table}
    \caption{\md{Surface reconstruction for sparse point clouds from PUGAN dataset where each shape only contains 512 points (Chamfer-L1$\times 10^3$).}}
    \vspace{-0.3cm}
    
    \centering
\resizebox{\linewidth}{!}{
\begin{tabular}{l|cc|cc}
\toprule
\multirow{1}*{} & \multicolumn{2}{c}{Chamfer-L1} & \multicolumn{2}{c}{F-Score}\\

Method  &Mean  &Median & $F1^{0.005}$ & $F1^{0.01}$\\ 
\midrule
ConvOcc \cite{peng2020convolutional} & 16.30 & 16.81 & 16.29 & 57.05 \\
POCO \cite{boulch2022poco} & 78.77 & 72.35 & 1.72 & 5.18 \\
\midrule
ConvOcc \cite{peng2020convolutional} + Prior & 13.73 & 14.02 & 18.60 & 66.03 \\
POCO \cite{boulch2022poco} + Prior & 10.29 &10.02 & 34.60 & 83.36 \\
Ours & \textbf{9.33} & \textbf{9.14} & \textbf{40.81} & \textbf{86.66} \\
\bottomrule
\end{tabular}}

    \label{table:upsamp_512}
\end{table}

\subsubsection{\md{Sparse point cloud surface reconstruction}}

\md{For evaluating the reconstruction performances from sparse point clouds, we conduct experiments under the widely-used PU-GAN \cite{li2019pu} dataset with extremely sparse point clouds containing only 256 or 512 points per shape. We introduce the pretrained APU-LDI \cite{li2024LDI} as the point cloud upsampling prior and reconstruct surfaces from the upsampled point clouds.}

\md{
In Tab. \ref{table:upsamp_256} and Tab. \ref{table:upsamp_512}, we show the quantitative comparison on reconstructing sparse point clouds with settings of 256 points and 512 points, respectively. The results demonstrate that the SoTA generalized methods ConvOcc and POCO struggle to produce robust reconstruction from the extremely sparse point clouds. By introducing expert point cloud upsampling methods as a prior to increase point densities, more accurate reconstructions can be achieved. CAP-UDF demonstrates superior performance compared to the generalized methods ConvOcc and POCO when utilizing the same upsampling prior, emphasizing its proficiency in faithfully reconstructing the point clouds enhanced by the prior. Please refer to Sec. 2.1 of the supplementary for visual comparisons on reconstructing sparse point clouds.
}

\subsection{\md{Large-Scale Scene Reconstruction}}
\md{Representing the surfaces for large point clouds using only one single neural network may struggle to produce high-fidelity reconstructions, due to the catastrophic forgetting of neural networks. To overcome this challenge, we introduce a slide-window strategy to reconstruct extremely large-scale scenes by splitting scenes into local chunks where each local chunk is represented with a specific neural network. The final reconstruction is achieved by fusing the local scenes.}

\md{We conduct experiments under the large-scale driving scenes from the KITTI \cite{geiger2012we} dataset. We use the LiDAR scans of frame 3000 to 4000 in Squeece00 subset of KITTI dataset pre-processed by NGS \cite{huang2022neural} as the full input, which are transformed into world coordinates using the provided camera trajectories. The results of directly reconstructing the full scene are shown in the ``Global Level'' of Fig. \ref{fig:kitti}, where only the overall outline of the scene is reconstructed and no geometric details are preserved, due to the catastrophic forgetting of neural networks, e.g., optimizing one local region results in degradation of other regions. 
}

\md{To overcome the limits of large-scale scene point cloud reconstruction with single neural network, we introduce a slide-window strategy to split the extremely large scene into local chunks at different scale levels and obtain the final scene reconstruction by fusing the reconstructed local geometries together. We reconstruct the scene under two local levels to split the full scene into 8 local trunks (Local Level-1) and 15 local trunks (Local Level-2). The size for local trunks in Local Level-2 is 51.2 $m^3$, and the trunks in Local Level-1 are achieved by merging two adjacent trunks in Local-Level-2. The results are shown in ``Local-Level 1 / Local-Level 2'' of Fig. \ref{fig:kitti}, where more subdivided local chunks lead to more detailed local geometries with higher qualities. We find that the Local Level-2 already produces convincing reconstructions, which can be considered as the "maximum" size before seeing performance degrade. And splitting scenes into more local trunks may cause space and time complexity for representing the full scene.
}

\begin{figure}[tb]
  \centering
  \includegraphics[width=1\columnwidth]{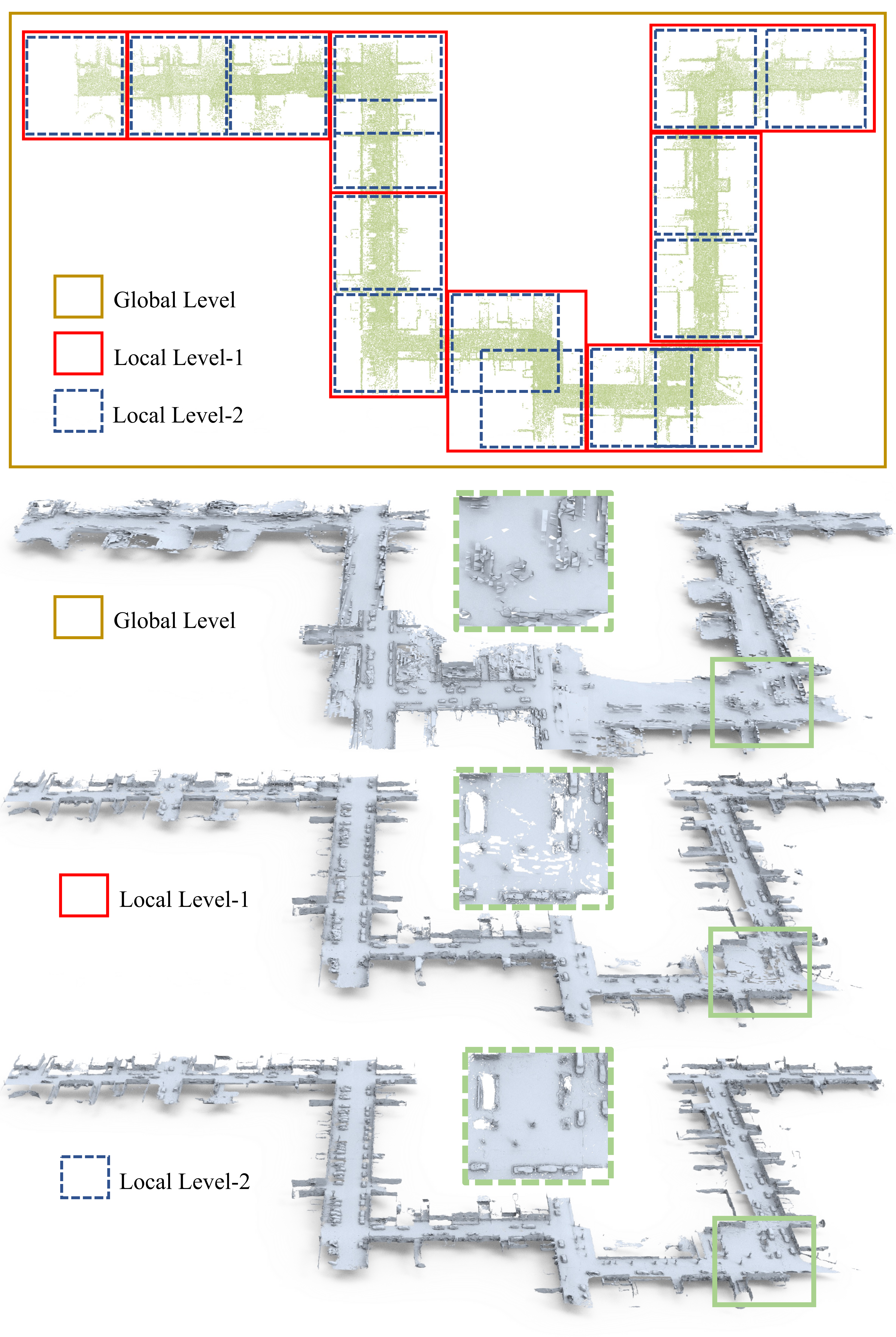}
  \caption{\md{Surface reconstruction results on the KITTI odometry dataset under different scale levels.}}
  \label{fig:kitti}
\end{figure}

\subsection{Ablation Study}
\label{section:sec4.4}
We conduct ablation studies to justify the effectiveness of each design in our method and the effect of some important parameters. We report the performance in terms of L2-CD under a subset of the ShapeNet Car dataset.
By default, all the experimental settings are kept the same as in Sec. \ref{section:sec4.1}, except for modified part described in each ablation experiment below.

\subsubsection{Framework design}
We first justify the effectiveness of each design of our framework in Tab. \ref{tab:ablation_settings}. We first directly use the loss proposed in Neural-Pull and find that the performance degenerates dramatically as shown by ``NP loss". We also use $g(x)=1-e^{(-x)}$ to replace $g(x)=|x|$ on the last layer of the network for $f$ before output, but found no improvement as shown by ``Exponent". We train the second stage from scratch as shown by ``Scratch" and prove that an end-to-end training strategy is more effective.

\begin{table}[h]
 \caption{Effect of framework design.}
\setlength{\tabcolsep}{2mm}
\resizebox{\linewidth}{!}{
 \begin{tabular}{c|c|c|c|c}
 \hline
 $\times 10^4$ & NP loss & Exponent & Scratch & Ours\\
 \hline
 L2CD & 0.2381 & 0.1218 & 0.1497 & \textbf{0.1112}\\
 \hline
 \end{tabular}}
 \label{tab:ablation_settings}
\end{table}

\subsubsection{The effect of stage numbers}
The number of stages during progressive surface approximation is also a crucial factor in the network training. We report the performance of training our network in different number of stages $St=[1,2,3,4]$ in Tab. \ref{tab:ablation_step}. We start the training of next stage after the previous one converges. We found that two stages training brings great improvement than training a single stage, and the improvements with 3rd and 4th stages are subtle. Therefore, we train CAP-UDF with two stages in practice.

\begin{table}[h]
 \caption{Effect of stage numbers.}
\setlength{\tabcolsep}{2.7mm}
\resizebox{\linewidth}{!}{
 \begin{tabular}{c|c|c|c|c}
 \hline
 $\times 10^4$ & 1 & 2 & 3 & 4\\
 \hline
 L2CD & 0.1218 & {0.1112} & 0.1107 & 0.1107\\
 \hline
 \end{tabular}}
 \vspace{-0.2cm}
 \label{tab:ablation_step}
\end{table}

\subsubsection{The effect of low confidence range}
We further explore the range of confidence region sample. Assume $\sigma$ as the range of high confidence region, we use 0.9$\sigma$, 1.0$\sigma$, 1.1$\sigma$ and 1.2$\sigma$ as the range of low confidence region. The results in Tab.  \ref{tab:ablation_scale} show that a too small or too large range will degenerate the performance.

\begin{table}[h]
 \caption{Effect of low confidence range.}
\setlength{\tabcolsep}{2.7mm}
\resizebox{\linewidth}{!}{
 \begin{tabular}{c|c|c|c|c}
 \hline
 $\times 10^4$ & 0.9 & 1.0 & 1.1 & 1.2\\
 \hline
 L2CD & 0.1133 & 0.1130 & \textbf{0.1112} & 0.1131 \\
 \hline
 \end{tabular}}
 \label{tab:ablation_scale}
\end{table}

\subsubsection{Surface extraction} We evaluate the effect of mesh refinement and the performance of different 3D grid resolutions. Tab. \ref{tab:ablation_mcubes} shows the accuracy and efficiency of different resolutions. We observe that the mesh refinement highly improves the accuracy and higher resolutions leads to better reconstructions at a cost of speed.

We provide the visualizations of the extracted surfaces with different resolutions as shown in Fig. \ref{fig:mcubes_exp}. It shows that a higher resolution leads to a more detailed reconstruction. Moreover, our designed mesh refinement operation brings great improvement in surface smoothness and local details due to the accurate unsigned distance values predicted by the neural network. 

\begin{table}[h]
\setlength{\tabcolsep}{2.5mm}
 \caption{Ablations on surface extraction.}
\resizebox{\linewidth}{!}{
 \begin{tabular}{c|c|c|c|c}
 \hline
 $\times 10^4$ & 64$^3$ & 128$^3$ & 256$^3$ & 320$^3$\\
 \hline
 w/o refine & 0.4169 & 0.1738 & 0.1294 & 0.1238 \\
 refine & 0.1606 & 0.1174 & \textbf{0.1112} & 0.1105 \\
 Time & 3.0 s & 21.9 s & 162.2 s & 307.6 s \\
 \hline
 \end{tabular}}
 \label{tab:ablation_mcubes}
\end{table}

\begin{figure}[tb]
  \centering
  \includegraphics[width=1\columnwidth]{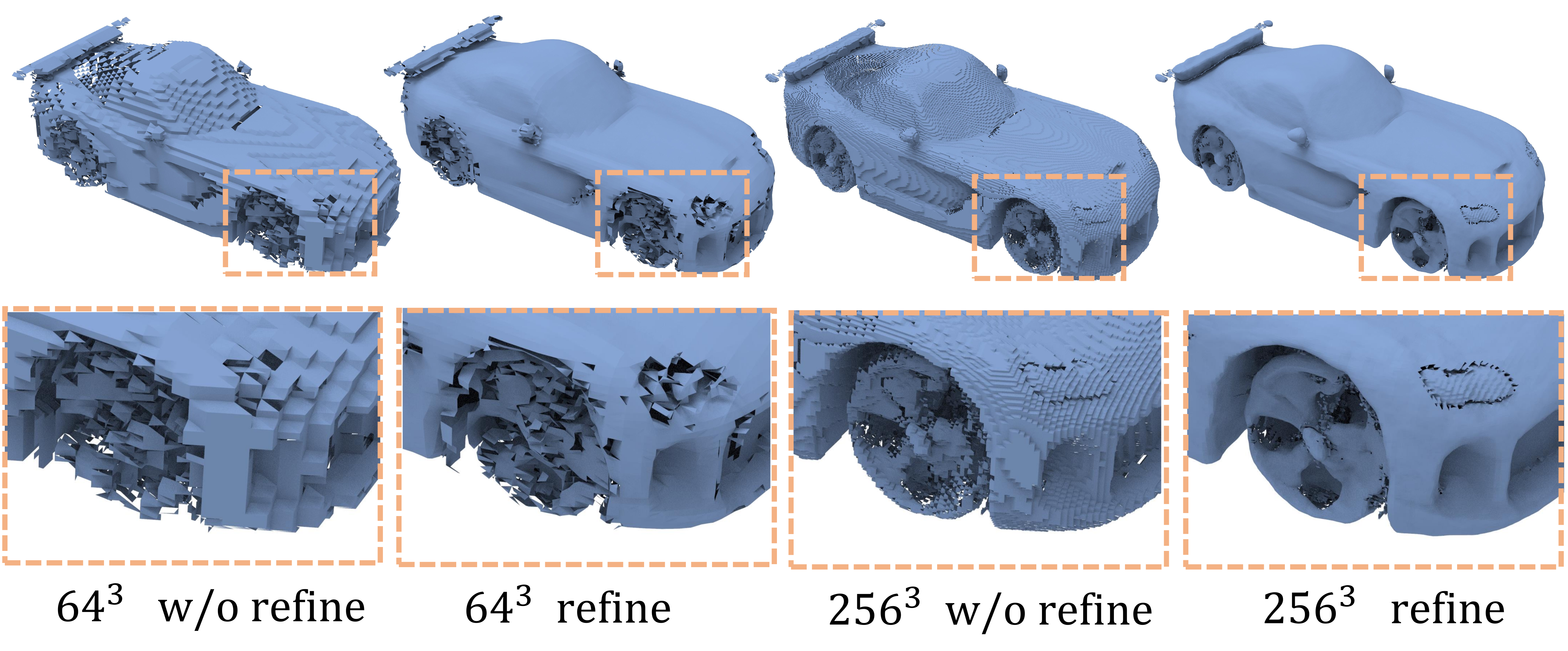}
  \caption{Visualizations of extracted mesh with different settings.}

  \label{fig:mcubes_exp}
\end{figure}

\subsubsection{The effect of training iterations}
We further test the effect of training iterations of the first and second stage. Since we use an end-to-end training strategy, we set a relatively small number of training iterations for the second stage. In experiments, we set the numbers of iteration in the first stage to 30k, 40k and 50k, and the second stage to 15k, 20k and 25k. For testing the effect of the iteration numbers in the first stage, we set the iteration numbers of the second stage to the default 20k, and the iteration numbers of the first stage is set to the default 40k while testing the second stage. The results in Tab. \ref{tab:train_iter} show that too few training iterations will lead to an under fitting problem and too many training iterations will result in network degeneration. 

\begin{table}[h]
 \caption{Effect on training iterations of different stages.}
\centering
\setlength{\tabcolsep}{2.7mm}
\resizebox{\linewidth}{!}{
 \begin{tabular}{c|ccc}
 \hline
 Stage1 & 30k & 40k & 50k\\
 \hline
 Chamfer-L2 & 0.1158 & \textbf{0.1112} & 0.1137\\
 \hline
 \hline
 Stage2 & 15k & 20k & 25k\\
 \hline
 Chamfer-L2 & 0.1125 & \textbf{0.1112} & 0.1133\\
 \hline
 \end{tabular}}
 \vspace{0.2cm}
 \vspace{-0.2cm}
 \label{tab:train_iter}
\end{table}

\subsubsection{The settings of normal estimation}
\label{section:normalablation}
We evaluate the effect of our designs for unsupervised point normal estimation in Tab. \ref{table:normal_ablation}. We first directly use the gradients at the location of raw point cloud as the estimated normal and find that the performance degenerates dramatically as shown by ``Point grad.''. The reason is that the normals are ambiguous at the zero-level set of UDFs, where we solve this problem by estimating the normals as the fusion of gradients at the queries sampled nearby. We use $K=[1,5,10,20,50,100]$ as the number of queries for estimating normals and find that a too small or too large $K$ will degenerate the performance.

\begin{table}[h]
\caption{Ablations on point normal estimation.}
\begin{center}
\setlength{\tabcolsep}{2.5mm}
\resizebox{\linewidth}{!}{
\begin{tabular}{c|l|ccc|c}
\hline\noalign{\smallskip}
~ & K \qquad & Clean & Strip & Gradient & average\\
\noalign{\smallskip}
\hline
\noalign{\smallskip}
\multirow{1}*{Point grad.} & 
 - & 7.30 & 7.74 & 7.78 & 7.61\\
\noalign{\smallskip}
\hline
\noalign{\smallskip}
\multirow{6}*{Query grad.} 
& 1 \qquad & 8.51 & 9.48 & 9.40 & 9.13\\
~ & 5 \qquad & 6.80 & 7.80 & 7.55 & 7.38\\
~ & 10 \qquad & 6.59 & 7.53 & 7.31 & 7.14\\
~ & 20 \qquad & 6.42 & 7.39 & 7.19 & 7.00\\
~ & 50 \qquad & 6.33 & \textbf{7.26} & \textbf{7.11} & \textbf{6.90}\\
~ & 100 \qquad & \textbf{6.32} & 7.27 & 7.14 & 6.91\\
\noalign{\smallskip}
\hline
\noalign{\smallskip}

\end{tabular}}
\end{center}
\label{table:normal_ablation}
\end{table}

\subsection{Efficiency Analysis}
\label{section:efficiency}
We analyse the efficiency of our proposed method by comparing the computational cost of field learning and mesh extraction with the state-of-the-art methods.

\subsubsection{Efficiency comparison on field learning}

We make a comparison with Neural-Pull \cite{ma2021neural}, IGR \cite{gropp2020implicit}, Point2mesh \cite{hanocka2020point2mesh} on the computational cost of optimizing for a single point cloud in Table \ref{table:computational}. The results show that our proposed method converges faster than all the baselines with fewer memory requirements.

\begin{table}[h]
     \caption{Efficiency comparison on field learning.}
    \centering
    \resizebox{1\linewidth}{!}{
    \begin{tabular}{l|cccc}
    \toprule
    
    Methods  &Neural-Pull & IGR & Point2mesh & Ours \\ 
    \midrule
    Time & 1150 s  & 1212 s  & 4028 s & \textbf{667 s} \\    
    Memory & 2.2 GB & 6.1 GB & 5.2 GB & \textbf{2.0 GB} \\
    \bottomrule
    \end{tabular}}
    \label{table:computational}
\end{table}

\subsubsection{Efficiency comparison on mesh extraction} 

We further evaluate the efficiency of our method. The comparison is shown in Tab. \ref{tab:comp_efficiency}. For reproducing the mesh generation process of NDF \cite{chibane2020neural}, we use the default 1$\times 10^6$ points and the parameters provided by NDF to generate a mesh with ball-pivoting-algorithm (BPA) \cite{bernardini1999ball}. It's clear that our method achieves a tremendous advantage over NDF even with a relatively high resolution (e.g.256$^3$). The reason is that our method allows a straightforward surface extracting from the learned UDFs while the ball-pivoting-algorithm used by NDF requires a lot of calculations for neighbour searching and normal estimation.

\begin{table}[h]
 \caption{Efficiency comparison of surface generation.}
\centering
\setlength{\tabcolsep}{1mm}
\resizebox{1\linewidth}{!}{
 \begin{tabular}{c|ccccc}
\toprule
 Method & NDF & Ours 64$^3$ & Ours 128$^3$ & Ours 256$^3$ & Ours 320$^3$\\
 \midrule
 Time & 2080.7 s & 3.0 s & 21.9 s & \textbf{162.2 s} & 307.6 s\\
 \bottomrule
 \end{tabular}}
 \vspace{0.2cm}
 \vspace{-0.2cm}
 \label{tab:comp_efficiency}
\end{table}

\subsection{More Visualizations and Analysis}
\label{section:more}
\begin{figure}[tb]
  \centering
  \includegraphics[width=\columnwidth]{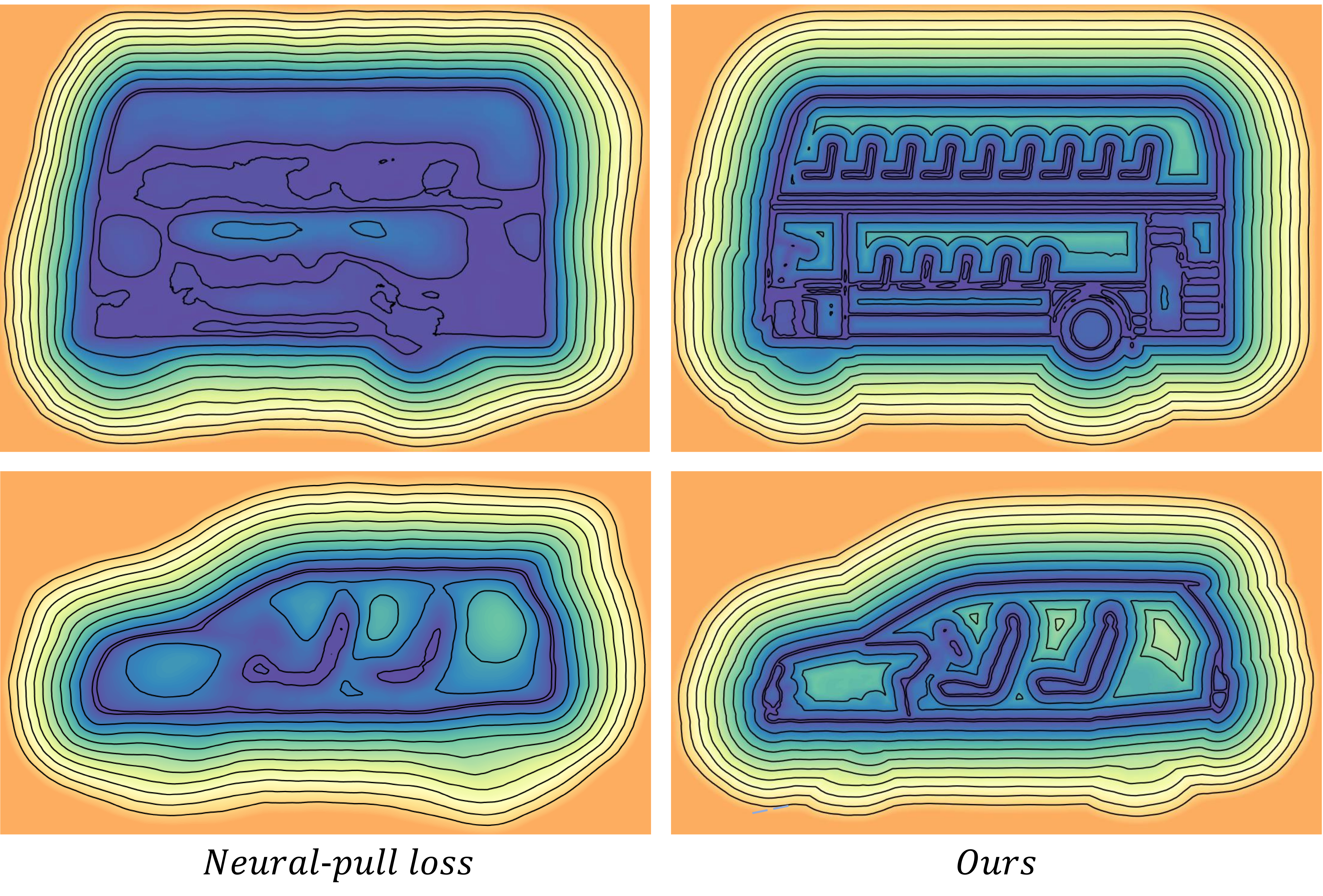}
  \caption{Visualizations of the unsigned distance field. The darker the color, the closer it is to the approximated surface. For a clear contrast, we set the color of the space far away from the approximated surface to orange.}
  \label{fig:supply_loss}
\end{figure}
\subsubsection{Visualizations of the unsigned distance field} 
To further evaluate the effectiveness of our consistency-aware field learning, we visualize the learned unsigned distance fields with Neural-Pull \cite{ma2021neural} loss in Eq. (\ref{eq:l2loss}) and our field consistency loss in Eq. (\ref{eq:gcloss}). Fig. \ref{fig:supply_loss} shows the visual comparison under two different shapes with complex inner structures. For the storey bus, training with the loss proposed by Neural-Pull \cite{ma2021neural} fails to handle the rich structures, thus leads to a chaotic distance field where no details were kept. On the contrary, our proposed consistency-aware learning can build a consistent distance field where the detailed structures are well preserved. It's clear that our method learns a highly continuous unsigned distance field and has the ability to keep the field around complex structures correct (e.g. the seats, tires and stairs), which also helps to extract a high fidelity surface.

\begin{figure}[h]
  \centering
  \includegraphics[width=1\columnwidth]{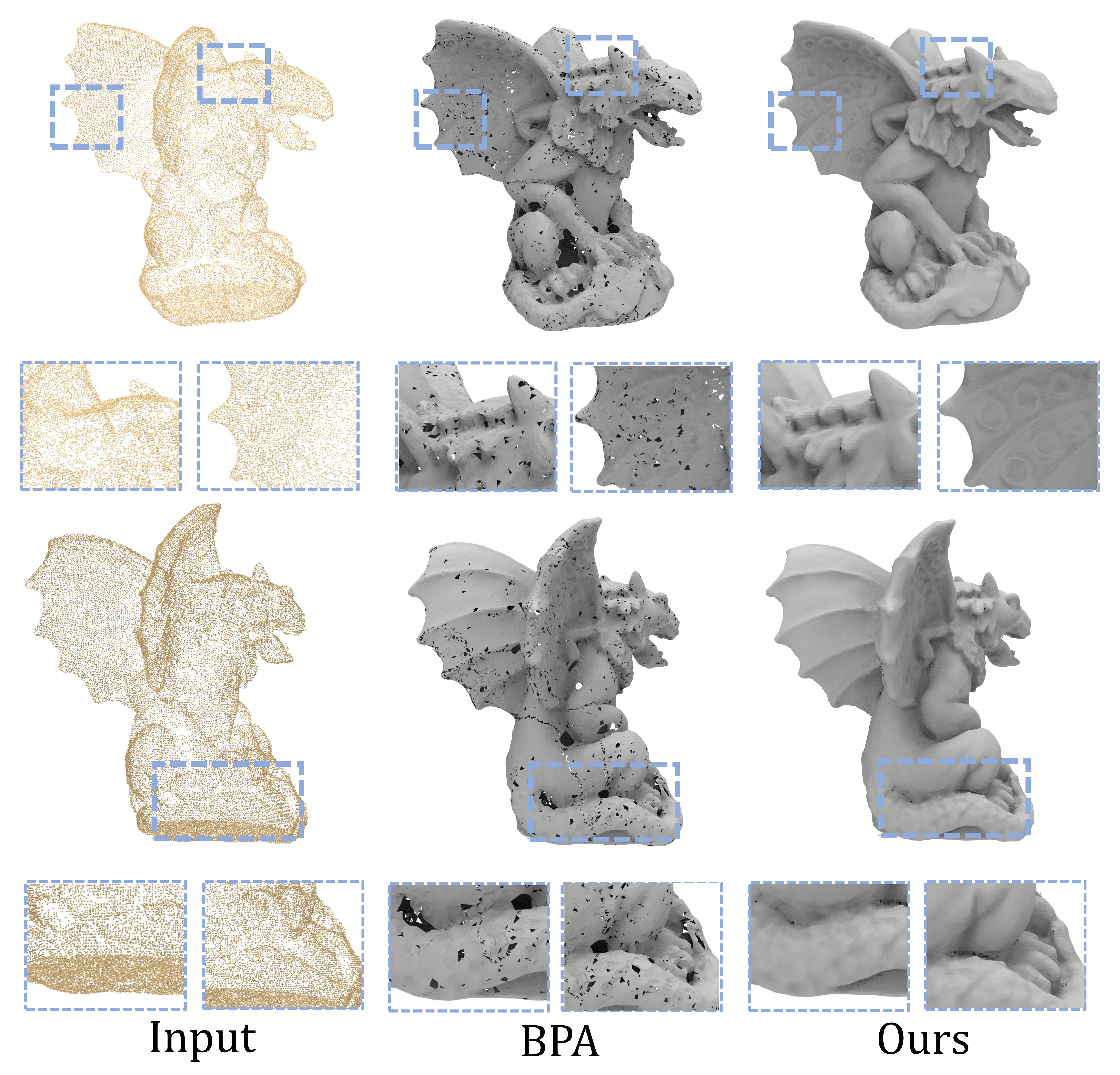}
  \caption{Visual comparison with ball-pivoting-algorithm.}
  \label{fig:supply_bpa}
\end{figure}

\subsubsection{Visual comparison with ball-pivoting-algorithm} 
To further explore the advantage of our straightforward surface extraction algorithm, we use the same setting as NDF \cite{chibane2020neural} to adopt ball-pivoting-algorithm (BPA) \cite{bernardini1999ball} to extract mesh from our generated point cloud and make a comparison with the mesh extracted using our method as shown in Fig. \ref{fig:supply_bpa}. Even with a carefully selected threshold, the mesh generated by BPA is still far from smooth and has a number of holes, and also fails to retain the detailed geometric information. On the contrary, our method allows to extract surfaces directly from the learned UDFs, thus is able to reconstruct a continuous and high-fidelity mesh where the geometry details is well preserved.

\section{Conclusion, Limitation and Future Work}
We propose a novel method to learn continuous UDFs directly from raw point clouds by learning to progressively move 3D queries onto the approximated surface. Our introduced reconstruction algorithm can extract surfaces directly from the gradient fields of the learned UDFs. Our method does not require ground truth distance values or point normals, and can reconstruct surfaces with arbitrary topology. We further extend our method to reconstruct surfaces from depth maps and estimate point normals without supervision, where we demonstrate our superior performance over the state-of-the-art in quantitative and qualitative results.

\md{
Finally, we acknowledge that there are some potential limitations to our method. First, directly leveraging CAP-UDF for reconstructing corrupted point clouds (e.g. occluded, noisy and sparse point clouds) may struggle to produce robust performances since we do not leverage any conditions or ground truth supervisions for training CAP-UDF. Although we provide an effective solution to improve the reconstruction quality by introducing priors from the state-of-the-art point cloud processing approaches in Sec. \ref{sec.4.6}, we believe there is room for further improvements, e.g., training CAP-UDF under diverse datasets in a data-driven way. Second, we use uniformly divided grids to extract surface, which can be improved with a coarse-to-fine paradigm. 
}

\appendices

\section{Proof of Surface Approximation Properties}
\label{section:proof}
We provide the proof of the assumption made for supporting progressive surface approximation in Sec. 3 of the main paper as follows.

\noindent\textbf{Assumption: } 
Given a raw point cloud which is a discrete representation of the surface, we have made a reasonable assumption: the closer the query location to the given point cloud, the smaller the error of searching the target point on the given point cloud. 

In the task of surface reconstruction, one of the key properties is local planarity, that is, the surface can be approximated by tangent plane with infinite accuracy. In graphics, the widely used surface representation is polygon mesh. To complete our proof, we provide a few necessary definitions.

\noindent\textbf{Definition 1.} Polygon mesh is a collection of vertices, edges and faces that defines the surface of 3D shape $O=( V , F )$. Vertices $V =\left\{v_{1}, \ldots, v_{r}\right\}$ determine the position of the faces in 3D space. The edges represent the connection between two vertices, thus a closed set of edges forming the faces $F =\left\{f_{1}, \ldots, f_{e}\right\}$.

\noindent\textbf{Definition 2.} Given a point cloud $P=\{p_i, i \in [1, N]\}$, its ground truth surface model $O=( V , F )$ consists of vertices $V =\left\{v_{1}, \ldots, v_{r}\right\}$ and faces $F =\left\{f_{1}, \ldots, f_{e}\right\}$. Given a set of query points $Q=\{q_i, i \in [1, M]\}$, the nearest point of query point $q_i$ on the model $O=( V , F )$ is $w_i$, and $w_i$ must be on one of the faces $F$. We define the meaning of error of searching for the target point on the given point cloud, $E(O,P,Q)=\sum_{i \in[1, M]}\min _{j \in[1, N]}\left\|w_{i}-p_{j}\right\|_{2}$.

Review our strategy for sampling query points: we sample $m$ queries around each point $p_i$ on $P$. A Gaussian function $\mathcal{N}(\mu, \sigma^2)$ is adopt to calculate the sampling probability for generating query location $q_{\sigma }^{i,j}, j \in [1,m]$, where $\mu=p_i$. Suppose we collect two sets of query points $Q_{\sigma_{1}}=\{ q_{\sigma_{1} }^{i,j} \mid i \in [1, N], j \in [1,m]\}$ and $Q_{\sigma_{2}}=\{ q_{\sigma_{2} }^{i,j} \mid i \in [1, N], j \in [1,m]\}$, where $\sigma_{1}$ and $\sigma_{2}$ are standard deviations of the sampled Gaussian function.

\noindent\textbf{Theorem 1.} Given a point cloud $P=\{p_i, i \in [1, N]\}$ and its ground truth surface model $O=( V , F )$. Assuming two sets of query points $Q_{\sigma_{1}}$ and $Q_{\sigma_{2}}$ are sampled, where $\sigma_{1} > \sigma_{2}$ , then $E(O,P, Q_{\sigma_{1}}) > E(O,P, Q_{\sigma_{2}})$.

\noindent\textbf{Proof 1.} Queries $q_{\sigma }^{i,j}, j \in [1,m]$ are sampled around point $p_i$ of the given point cloud $P$. Assume that the nearest surface point on the model $O=( V , F )$ to the query point $q_{\sigma }^{i,j}$ is $w_{\sigma }^{i,j}$, where $w_{\sigma }^{i,j}$ is on the face $ f_{\sigma }^{i,j,k}$. The point $p_i$ must be on one of the faces $F$, assuming that this face is $f_{u}$. Let $Q_{\sigma_{1}}^{i}=\left\{q_{\sigma_{1} }^{i,j} \mid j \in [1,m]\right \}$,$Q_{\sigma_{2}}^{i}=\left\{q_{\sigma_{2} }^{i,j} \mid j \in [1,m]\right \}$, where $\sigma_{1} > \sigma_{2}$. \md{We decompose $Q_{\sigma_{1}}^{i}$ and $Q_{\sigma_{2}}^{i}$ to the two sides: (1) The set of queries where their nearest surface points $w_{\sigma }^{i,j}$ lie on the face $f_u$, indicated as: $Q_{\sigma_{1},i}^{+}=\left \{ {q_{\sigma_{1} }^{i,j} \mid f_{\sigma_{1}}^{i,j,k}=f_{u}} \right \}$,$Q_{\sigma_{2},i}^{+}=\left \{ {q_{\sigma_{2} }^{i,j} \mid f_{\sigma_{2}}^{i,j,k}=f_{u}} \right \}$, and (2) the set of queries where their nearest surface points $w_{\sigma }^{i,j}$ lie outside the face $f_u$, indicated as: $Q_{\sigma_{1},i}^{-}=\left \{ {q_{\sigma_{1} }^{i,j} \mid f_{\sigma_{1}}^{i,j,k} \ne f_{u}} \right \}$, $Q_{\sigma_{2},i}^{-}=\left \{ {q_{\sigma_{2} }^{i,j} \mid f_{\sigma_{2}}^{i,j,k} \ne f_{u}} \right \}$. Therefore $E(O,P, Q_{\sigma_{1}}^{i})=E(O,P, Q_{\sigma_{1},i}^{+})+E(O,P, Q_{\sigma_{1},i}^{-})$, $E(O,P, Q_{\sigma_{2}}^{i})=E(O,P, Q_{\sigma_{2},i}^{+})+E(O,P, Q_{\sigma_{2},i}^{-})$. }

\md{The sampling probability is calculated based on Gaussian function, that is defined by density function $ \mu(q)$, and assuming that $f_{u}=n^{T}x+d$, where $n$, $x$ and $d$ are the normal at $f_{u}$, the 3D location and the offset to the coordinate origin. The error at $Q_{\sigma_{1},i}^{+}$ is computed as:}

\begin{equation}\small
\begin{split}
E(O,P, Q_{\sigma_{1},i}^{+})=\int ||w_{ \sigma_{1} }^{i,j} &-p_{i}\|_{2}\mu(q_{\sigma_{1} }^{i,j}) d(q_{\sigma_{1} }^{i,j})\\
= \sum_{j \in[1, m]}&||q_{\sigma_{1} }^{i,j}-\frac{q_{\sigma_{1} }^{i,j} \cdot n^{T}+d}{n^{T} \cdot n^{T}} \cdot n^{T}-p_{i}\|_{2}\\
&\cdot \frac{1}{\sqrt{(2 \pi)^{3}\sigma_{1}|I|}} e ^{-\frac{1}{2}( q_{\sigma_{1} }^{i,j} - p_{i} )^{ T } \sigma_{1}I( q_{\sigma_{1} }^{i,j} - p_{i} )}
\end{split}
\end{equation}


Since $E(O,P, Q_{\sigma_{1},i}^{+})$ will decrease with decreasing $\sigma_{1}$, then we get  $E(O,P, Q_{\sigma_{1},i}^{+})> E(O,P, Q_{\sigma_{2},i}^{+})$.

The geometric distribution of $f_{\sigma}^{i,j,k}$ appearing around $f_{u}$ can be regarded as randomly distributed, the $E(O,P, Q_{\sigma,i}^{-})$ is also randomly distributed and independent of the density function, so that in a statistical sense, $ E(O,P, Q_{\sigma{1},i}^{-})= E(O,P, Q_{\sigma_{2},i}^{-})$. Then we get  $E(O,P, Q_{\sigma_{1},i}) > E(O,P, Q_{\sigma_{2},i})$. The case of arbitrary other $Q_{\sigma_{1}}^{j}$ and $Q_{\sigma_{2}}^j$ can be proved similarly. \md{So we prove} $E(O,P, Q_{\sigma_{1}}) > E(O,P, Q_{\sigma_{2}})$.

\section{Additional Experiments}

\begin{figure}[tb]
  \centering
  \includegraphics[width=1\columnwidth]{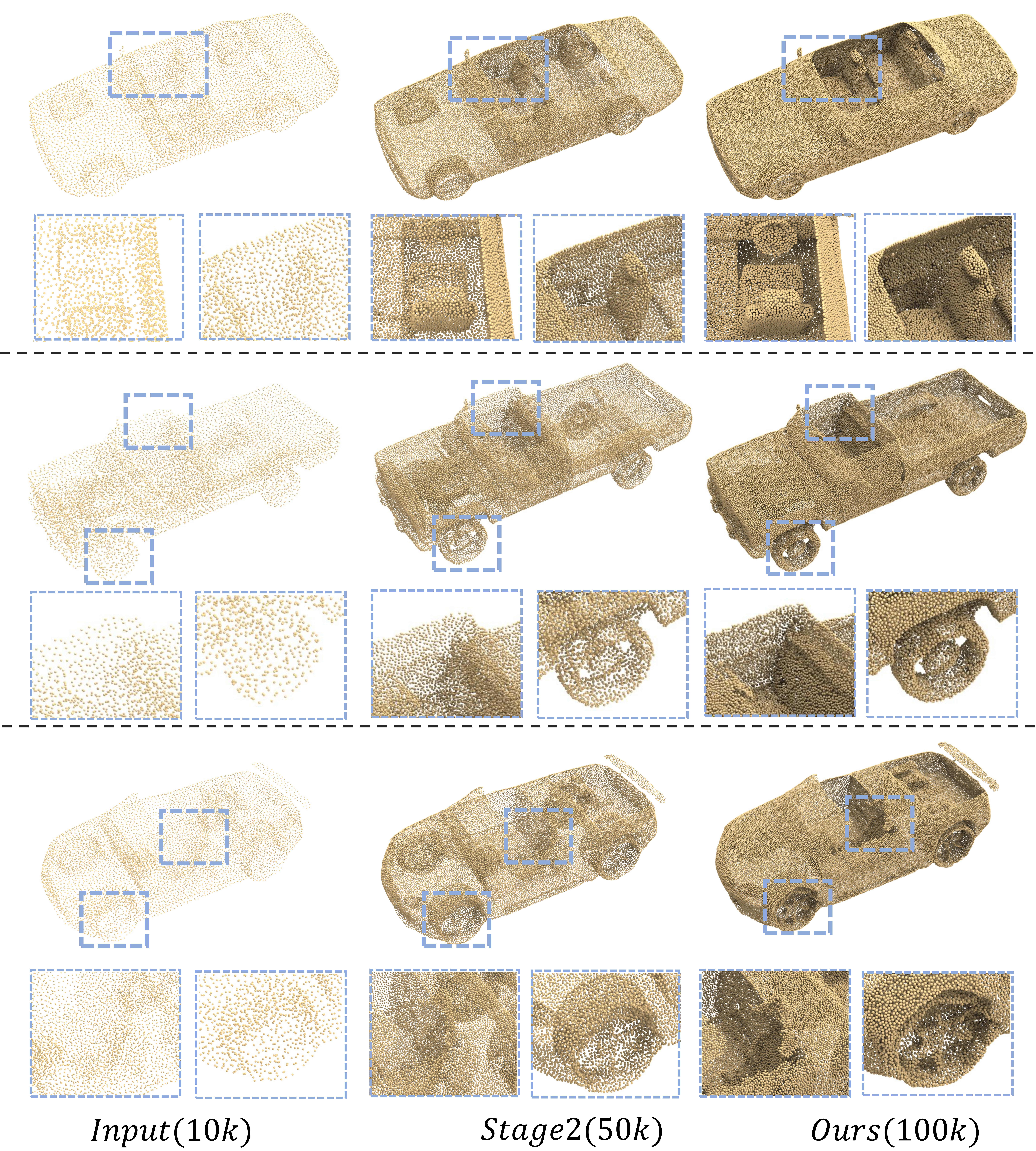}
  \caption{Visualizations of our generated intermediate and final point clouds.}
  \label{fig:multistep_exp}
\end{figure}

\subsection{\md{Visual Comparisons on Reconstructing Corrupted Point Clouds}}\md{We provide more visual comparisons on reconstructing corrupted point clouds in Fig. \ref{fig:prior_comp}. We compare our proposed CAP-UDF with the SOTA generalized surface reconstruction methods, e.g., ConvOcc \cite{peng2020convolutional} and POCO \cite{boulch2022poco}. }

\md{The visual comparisons on reconstructing occluded, noisy and sparse point clouds is shown in (a), (b) and (c) in Fig. \ref{fig:prior_comp}. The results demonstrate that the generalized methods fail to produce robust reconstruction with complete and clean geometries from the corrupted point clouds. We introduce expert point cloud processing models as priors to first improve the quality of point clouds before
learning distance fields for reconstruction, which enables CAP-UDF for robust reconstructions. CAP-UDF also demonstrates superior performance in faithfully reconstructing improved point clouds when compared to the generalized methods ConvOcc  and POCO, all using the same point cloud processing priors.}
 
\begin{figure*}[!tb]
  \centering
  \includegraphics[width=2\columnwidth]{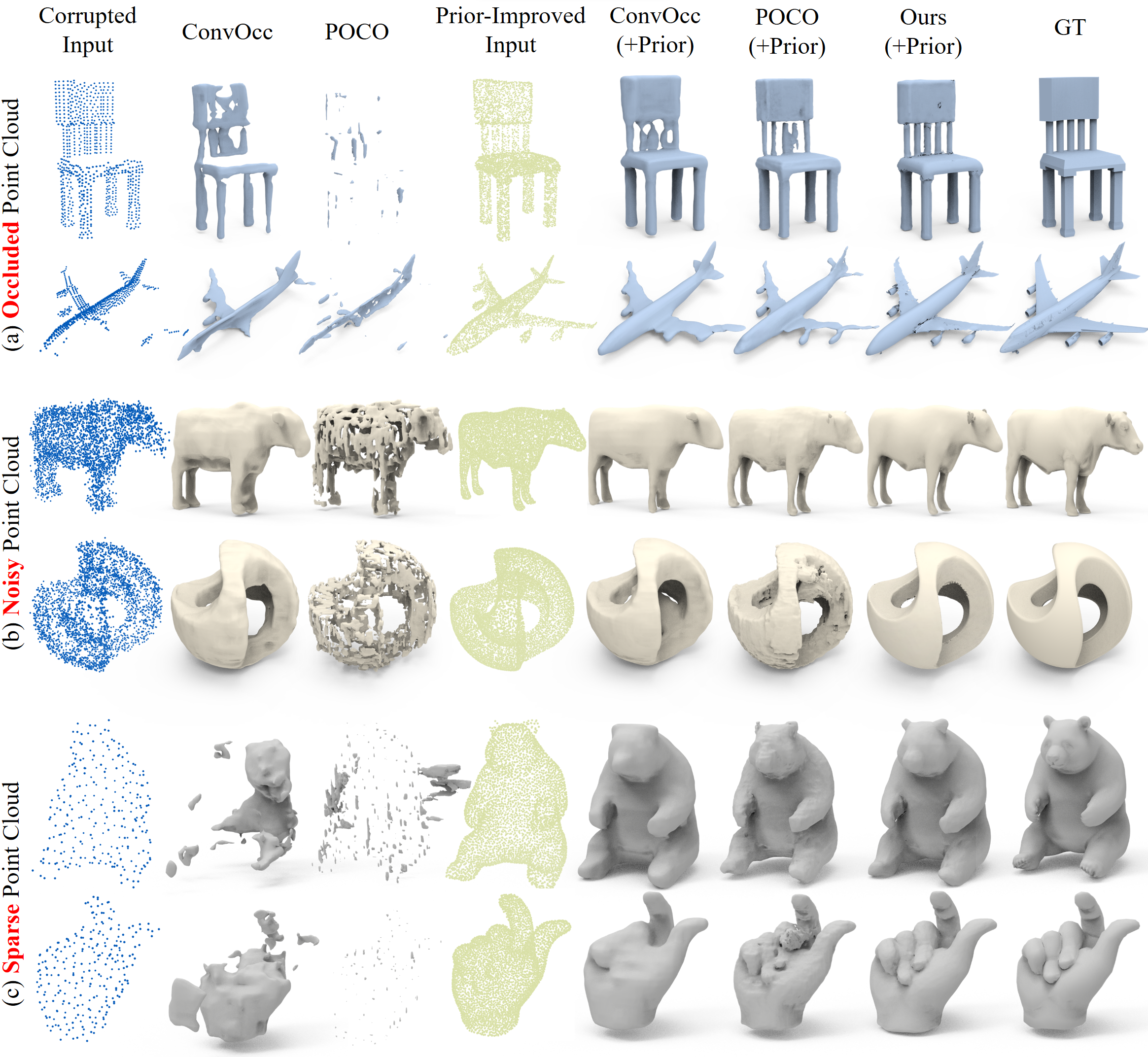}
  \caption{\md{Visual comparisons of surface reconstruction from corrupted point clouds. `Prior-Improved Input' indicates the processed point clouds with expert prior models for each corrupted type. `+Prior' indicates reconstructions with prior-improved inputs. (a) Reconstructing occluded point clouds, the expert prior model is AdaPoinTr \cite{yu2023adapointr}. (b) Reconstructing noisy point clouds, the expert prior model is Iterative-PFN \cite{de2023iterativepfn}. (c) Reconstructing sparse  point clouds (i.e. 256 points per shape), the expert prior model is APU-LDI \cite{li2024LDI}.}}
  \label{fig:prior_comp}
\end{figure*}

\subsection{Complete Comparisons under 3D Scene Dataset}
We also provide the complete comparison under all the five scenes of the 3D Scene dataset \cite{zhou2013dense}. The results is shown in Tab. \ref{table:scenes_supp}. We provide a comprehensive comparison with NDF by comparing both the generated point cloud (*$_{PC}$) and the generated mesh (*$_{mesh}$). We use L1 and L2 chamfer distance (L2CD, L1CD) as evaluation metrics.

\begin{table*}[h]\scriptsize

\centering
   \caption{Surface reconstruction for point clouds under 3D Scene (L2CD$\times 1000$).}
\centering

\resizebox{\linewidth}{!}{
    \begin{tabular}{c|c|c|c||c|c||c|c||c|c||c|c}
     \hline

        &&\multicolumn{2}{c||}{Burghers}&\multicolumn{2}{c||}{Lounge}&\multicolumn{2}{c||}{Copyroom}&\multicolumn{2}{c||}{Stonewall}&\multicolumn{2}{c}{Totempole}\\
        \hline
        &&L2CD&L1CD&L2CD&L1CD&L2CD&L1CD&L2CD&L1CD&L2CD&L1CD\\
     \hline
     \multirow{6}{*}{\rotatebox{90}{100/$m^2$}}&COcc~\cite{peng2020convolutional}&8.904&0.040&6.979 &0.041 &6.78 &0.041 &12.22 &0.051 &4.412 &0.041\\
     &LIG~\cite{jiang2020local}&3.112 &0.044 &9.128 &0.054 &4.363 &0.039 &5.143 &0.046 &9.58 &0.062\\   
     &DeepLS~\cite{chabra2020deep}&3.111&0.050&3.894&0.056&1.498&0.033&2.427&0.038 &4.214&0.043 \\    
     &NDF$_{PC}$ \cite{chibane2020neural}&0.320&0.012 &0.417&0.013 &0.291&0.012 &0.252&0.010 &0.767&0.015 \\
     &NDF$_{mesh}$ \cite{chibane2020neural}&0.463&0.014&0.484&0.015&0.439&0.015&0.248&0.011&0.624&0.014\\
     &OnSurf \cite{On-SurfacePriors}&0.544&0.018 &0.435&0.013 &0.434&0.017 &0.371&0.016 &3.986&0.040 \\
     \cline{2-12}
     &Ours$_{PC}$&\textbf{0.121}&\textbf{0.010}&\textbf{0.277}&\textbf{0.013}&\textbf{0.150}&\textbf{0.010}&\textbf{0.079}&\textbf{0.008}&\textbf{0.090}&\textbf{0.008}\\
     &Ours$_{mesh}$&\textbf{0.212}&\textbf{0.011}&\textbf{0.245}&\textbf{0.011}&\textbf{0.214}&\textbf{0.012}&\textbf{0.118}&\textbf{0.009}&\textbf{0.145}&\textbf{0.009}\\
     
     \hline
     \hline
     \multirow{6}{*}{\rotatebox{90}{500/$m^2$}}&COcc~\cite{peng2020convolutional}&26.97&0.081 &9.044 &0.046  &10.08 &0.046  &17.70 &0.063  &2.165 &0.024  \\
     &LIG~\cite{jiang2020local}&3.080 &0.046  &6.729 &0.052  &4.058 &0.038  &4.919 &0.043   &9.38 &0.062  \\
     &DeepLS~\cite{chabra2020deep}&0.714&0.020 &10.88&0.077 &0.552&0.015 &0.673&0.018 &21.15&0.122 \\
     &NDF$_{PC}$ \cite{chibane2020neural}&0.304&0.014 &0.261&0.011 &0.162&0.010 &0.239&0.012 &0.919&0.021 \\
     &NDF$_{mesh}$ \cite{chibane2020neural}&0.546&0.018&0.314&0.012&0.242&0.012&0.226&0.012&1.049&0.025\\
     &OnSurf \cite{On-SurfacePriors}&0.609&0.018 &0.529&0.013 &0.483&0.014 &0.666&0.013 &2.025&0.041 \\
     \cline{2-12}
     &Ours$_{PC}$ &\textbf{0.072}&\textbf{0.008}&\textbf{0.146}&\textbf{0.011}&\textbf{0.072}&\textbf{0.008}&\textbf{0.038}&\textbf{0.007}&\textbf{0.064}&\textbf{0.008}\\   
     &Ours$_{mesh}$ &\textbf{0.192}&\textbf{0.011}&\textbf{0.099}&\textbf{0.009}&\textbf{0.120}&\textbf{0.009}&\textbf{0.069}&\textbf{0.008}&\textbf{0.131}&\textbf{0.010}\\
     
     \hline
     \hline
     \multirow{6}{*}{\rotatebox{90}{1000/$m^2$}}&COcc~\cite{peng2020convolutional}&27.46&0.079 &9.54 &0.046  &10.97 &0.045  &20.46 &0.069  &2.054 &0.021 \\
     &LIG~\cite{jiang2020local}&3.055 &0.045  &9.672 &0.056  &3.61 &0.036  &5.032 &0.042  &9.58 &0.062  \\
    &DeepLS~\cite{chabra2020deep}&0.401&0.017 &6.103&0.053 &0.609&0.021 &0.320&0.015 &0.601&0.017 \\
     &NDF$_{PC}$ \cite{chibane2020neural}&0.575&0.019 &0.303&0.012 &0.186&0.011 &0.407&0.016 &1.333&0.026 \\

     &NDF$_{mesh}$ \cite{chibane2020neural}&1.168&0.027&0.393&0.014&0.269&0.013&0.509&0.019&2.020&0.036\\
     &OnSurf \cite{On-SurfacePriors}&1.339&0.031 &0.432&0.014 &0.405&0.014 &0.266&0.014 &1.089&0.029 \\
    \cline{2-12}
    &Ours$_{PC}$&\textbf{0.087}&\textbf{0.010}&\textbf{0.057}&\textbf{0.009}&\textbf{0.083}&\textbf{0.010}&\textbf{0.043}&\textbf{0.007}&\textbf{0.086}&\textbf{0.010}\\   
    &Ours$_{mesh}$&\textbf{0.191}&\textbf{0.010}&\textbf{0.092}&\textbf{0.008}&\textbf{0.113}&\textbf{0.009}&\textbf{0.066}&\textbf{0.007}&\textbf{0.139}&\textbf{0.009}\\
   \hline
   \end{tabular}}

   \label{table:scenes_supp}
\end{table*}

\subsection{Visualization of our generated intermediate and final point clouds}
As shown in Fig. \ref{fig:multistep_exp}, we provide the visualizations of the raw input point cloud as `Input' and the intermediate point cloud updated by the well-moved queries which serves as the target point cloud at the second stage, as shown by `Stage2'. We also provide the generated dense point cloud by moving the queries to the approximated surface using Eq. (1) of the main paper, as shown by `Ours'. The raw input point cloud is highly discrete and fails to provide detailed supervision for surface reconstruction. On the contrary, the intermediate point cloud updated by the well-moved queries shows great uniformity and continuity, thus can provide the supervision of fine geometries such as the detailed steering wheel and tires of the car. The quantitative comparison of using the raw input point cloud and the intermediate point cloud as target for surface reconstruction is shown in Tab. 13 of the main paper, where leveraging the intermediate point cloud as target can achieve $10\%$ improvement over leveraging the raw input point cloud. Furthermore, due to our progressive surface approximation strategy, our generated dense point cloud can maintain uniform and keep the detailed geometric.

\subsection{More Visualization Comparisons}
We provide more visual comparisons under Surface Reconstruction Benchmark (SRB) dataset \cite{williams2019deep} in Fig. \ref{fig:sup_spb} and MGD dataset \cite{bhatnagar2019multi} in Fig. \ref{fig:mgd_supp}, respectively.

\begin{figure*}[!t]
    \centering
    \includegraphics[width=2\columnwidth]{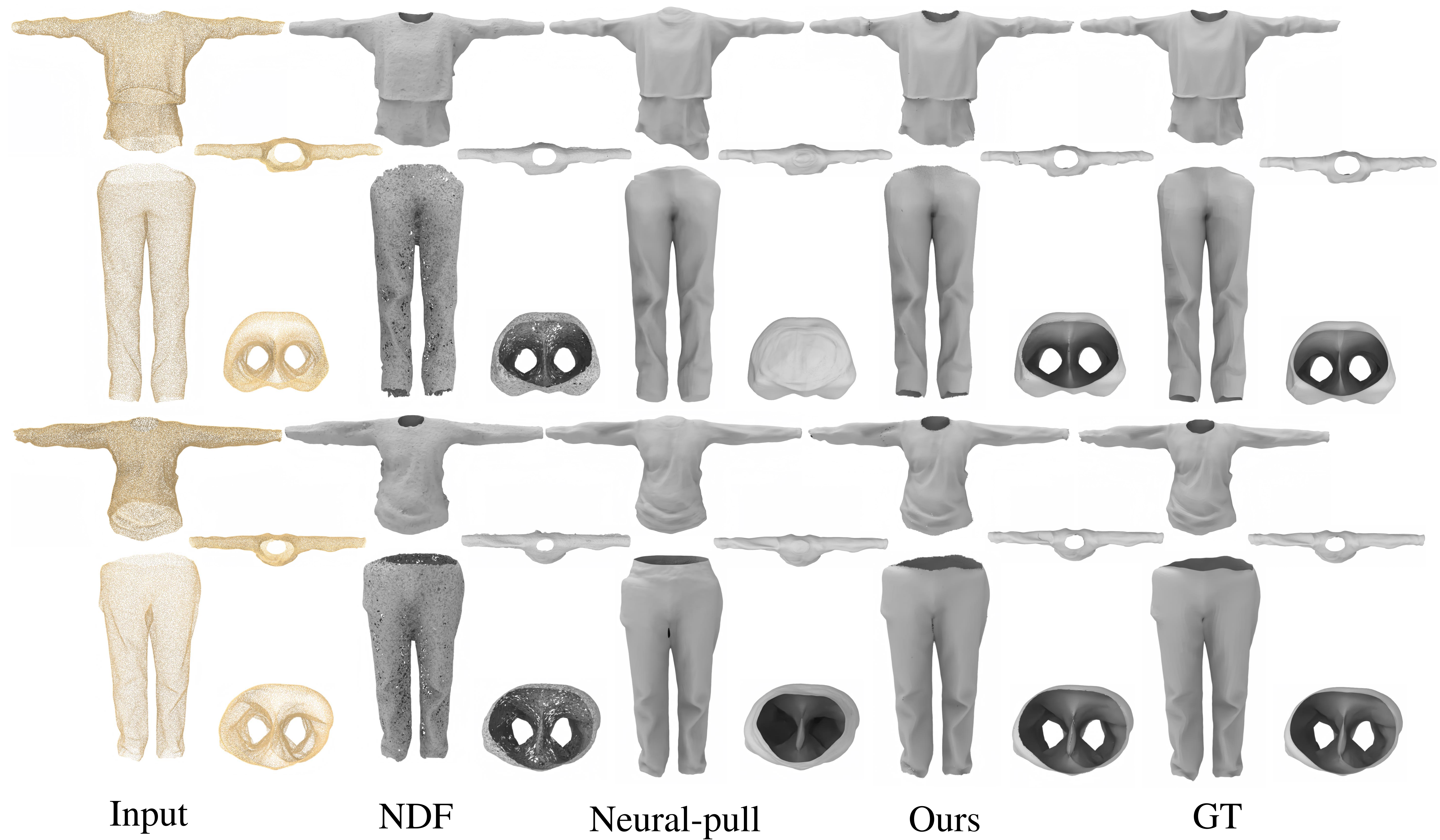}
    \caption{Visual comparisons of surface reconstruction on MGN dataset.}
    \label{fig:mgd_supp}
\end{figure*}

\begin{figure*}[tb]
  \centering
  \includegraphics[width=2\columnwidth]{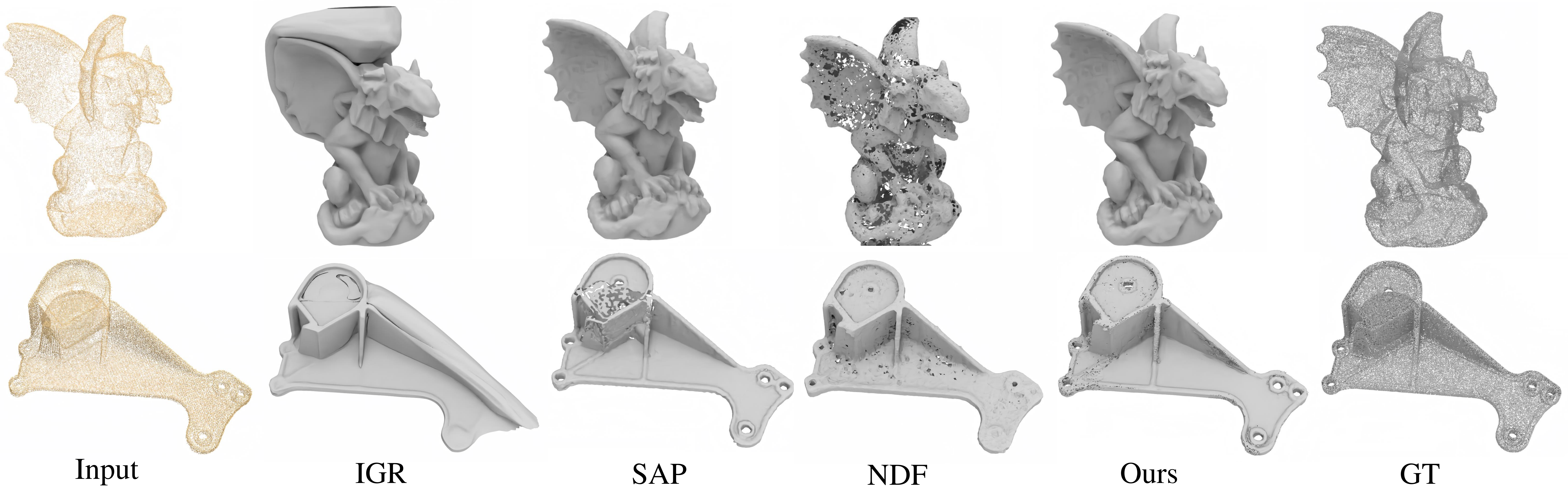}
  \caption{More visualizations under SRB dataset. }
  \label{fig:sup_spb}
  \vspace{-0.5cm}
\end{figure*}

\bibliographystyle{IEEEtran}
\bibliography{egbib}

\end{document}